\documentclass{article}

     \usepackage[preprint]{neurips_2019}

\usepackage[utf8]{inputenc} 
\usepackage[T1]{fontenc}    
\usepackage{hyperref}       
\usepackage{url}            
\usepackage{booktabs}       
\usepackage{amsmath,amsfonts,amssymb,amsthm}
\usepackage{nicefrac}       
\usepackage{microtype}      
\usepackage{algorithm}
\usepackage{algorithmic}
\usepackage{graphicx}
\usepackage{color}
\usepackage{natbib}

\usepackage{soul}
\soulregister\cite7 
\soulregister\citep7 
\soulregister\citet7 
\soulregister\ref7 

\DeclareMathOperator*{\argmax}{arg\,max}

\theoremstyle{plain}
\newtheorem{assumption}{Assumption}
\newtheorem{proposition}{Proposition}
\newtheorem{theorem}{Theorem}
\newtheorem*{remark}{Remark}
\newtheorem{lemma}{Lemma}

\title{A Dimension-free Algorithm for Contextual Continuum-armed Bandits}

\author{
	Wenhao Li \\
	Department of Management Sciences \\
	City University of Hong Kong \\
	\texttt{wenhaoli6-c@my.cityu.edu.hk} \\
	\And
	Ningyuan Chen \\
	Rotman School of Management \\
	University of Toronto\\
	\texttt{Ningyuan.chen@rotman.utoronto.ca} \\
	\And
	L.Jeff Hong \\ 
	School of Management and School of Data Science \\
	Fudan University \\
	\texttt{hong\_liu@fudan.edu.cn} \\
}

\begin{document}

\maketitle

\begin{abstract}
In contextual continuum-armed bandits, the contexts $x$ and the arms $y$ are both continuous and drawn from high-dimensional spaces.
The payoff function to learn $f(x,y)$ does not have a particular parametric form.
The literature has shown that for Lipschitz-continuous functions, the optimal regret is $\tilde{O}(T^{\frac{d_x+d_y+1}{d_x+d_y+2}})$, where $d_x$ and $d_y$ are the dimensions of contexts and arms, and thus suffers from the curse of dimensionality.
We develop an algorithm that achieves regret $\tilde{O}(T^{\frac{d_x+1}{d_x+2}})$ when $f$ is globally concave in $y$.
The global concavity is a common assumption in many applications.
The algorithm is based on stochastic approximation and estimates the gradient information in an online fashion. {Furthermore, we generalize our algorithm to an adaptive scheme that takes advantage of sequentially arrived context without sacrificing the worst-case performance.} Our results generate a valuable insight that the curse of dimensionality of the arms can be overcome with some mild structures of the payoff function.
\end{abstract}
\section{Introduction}\label{sec:introduction}
The multi-armed bandits (MAB) deal with a class of sequential decision making problems \citep{lai1985asymptotically,auer2002finite,bubeck2012regret}.
Without knowing the payoff of each decision, the decision maker chooses a decision from a set of alternatives (arms) in each epoch based on the past history.
The observed random payoff associated with the chosen decision can be used to learn in future epochs.
The goal is to maximize the total payoff over a finite horizon.
The MAB setting has been introduced in \citet{robbins1952some} and studied intensively since then in statistics, computer sciences, operations research, and economics.

Recently, the contextual bandits problems have received attentions of many scholars. Before making decision in each epoch, the decision maker receives a context that can be used to infer the payoff and suggest which arm to pull.
Contextual bandits are motivated by advertisement placement on webpages.
Upon observing the user profile (context), the firm needs to decide which advertisement to place (arm) that may interest the user.
The number of clicks is the payoff to maximize.

In this paper, we consider contextual continuum-armed bandits.
Compared to the classic MAB setting, both decision and context are drawn from continuous spaces in our problem.
This is motivated by personalized pricing in operations research and marketing.
A firm sells multiple products over a finite selling season via dynamically adjusted prices.
For each coming customer, the firm observes her profile such as education background, zip code, age and purchasing history.
Then the firm decides a personalized price vector for the customer.
The total payoff is the revenue extracted from a finite number of customers.
The continuous nature and high dimensionality of the customer profile and pricing motivate the contextual continuum-armed bandit problem.


\textbf{Prior Work.} There is a body of literature on continuum-armed bandits \citep{agrawal1995continuum, pandey2007bandits, kleinberg2010sharp, maillard2010online}.
\citet{kleinberg2005nearly} studies the case that the mean payoff function satisfies a H\"{o}lder continuous property with constant $\alpha$. This work proves a worst-case lower bound  $O(T^{\frac{\alpha+1}{2 \alpha +1}})$ for the regret of any algorithm when the arms set is one dimensional and proposes a uniform discretization algorithm achieving a regret of order $\tilde{O}(T^{\frac{\alpha+1}{2 \alpha +1}})$, nearly tight with the lower bound. \citet{auer2007improved} also study the one-dimensional arm set. Under the condition that payoff functions have finitely many maxima and a non-vanishing, continuous second-order derivative at all maxima, their algorithm achieves the regret order $\tilde{O}(\sqrt{T})$. 
\citet{kleinberg2008multi} consider the multi-dimensional case and generalize the Lipschitz bandit problem to metric space. They present an algorithm obtaining the regret $\tilde{O} (T^{\frac{d+1}{d+2}})$ where $d$ is the covering dimension of arm space, kindly capturing the sparsity of arm space. The same regret is achieved by \citet{bubeck2011x}, but $d$ is the packing dimension instead. They further demonstrate that the smoothness of the mean payoff function around its maximum can be used to reduce the packing dimension.
Regret bounds independent of the dimension of the arm space are obtained by \citet{cope2009regret,agarwal2013stochastic}. \citet{cope2009regret} shows an asymptotic regret bound of size $O(\sqrt{T})$ if the payoff functions are unimodal, three times continuously differentiable and its derivative is well behaved at its maximal. \citet{agarwal2013stochastic} assume globally convex and Lipschitz payoff functions and achieving a regret $\tilde{O} (poly(d) \sqrt{T})$ with high probability.

Another stream of related literature is contextual bandits. \citet{woodroofe1979,Wang2005Bandit,rigollet2010nonparametric,Perchet2013The} study contextual MAB with stochastic payoffs, under the name \textit{bandits with covariates}: the context is a random variable correlated with the payoffs. They consider the case of finitely discrete arms. On the other hand, \citet{slivkins2014contextual, lu2009showing} consider continuous arms and assume Lipschitz continuity for both the arm and context space.
They prove a lower bound $O(T^{\frac{\hat{d}_x+\hat{d}_y+1}{\hat{d}_x+\hat{d}_y+2}})$ for the regret of any algorithm where $\hat{d}_x,\hat{d}_y$ are the packing dimensions of the context and arms space respectively. \citet{lu2009showing} presents a uniformly partition algorithm obtaining nearly tight regret upper bound $\tilde{O}(T^{\frac{d_x+d_y+1}{d_x+d_y+2}})$ where $d_x,d_y$ are covering dimensions of the context and arms space. The same regret bound can be achieved by the uniform partition and a bandit-with-expert-advice algorithm such as EXP4 \citep{auer2002nonstochastic} or NEXP \citep{mcmahan2009tighter}. The uniform partition is used to define an expert whose advise is simply an arbitrary arm for each set of the partition. \citet{slivkins2014contextual} proposes an adaptive zooming algorithm so that frequently occurring contexts and high-paying arms structure can be used to improve practical performance.  There other versions of contextual bandit problems, such as linear bandits \citep{auer2002using,dani2008stochastic,rusmevichientong2010linearly,abbasi2011improved}, contextual bandits with policy sets \citep{auer2002nonstochastic,langford2008epoch, agarwal2012contextual, dudik2011efficient}.

In operations research, many papers have focused on dynamic pricing and demand learning \citep{besbes2009dynamic,besbes2015surprising,broder2012dynamic,den2013simultaneously,den2015dynamic}.
Recently \citet{chen2018nonparametric} consider personalized pricing of a single product to customers.
Besides Lipschitz continuity in arms and context space, they further assume smoothness and local concavity at the unique maximizer of the payoff function.
Their algorithm achieves the near-optimal regret in their setting, $\tilde{O}(T^{\frac{d_x+2}{d_x+4}})$, slightly better than the $\tilde{O}(T^{\frac{d_x+d_y+1}{d_x+d_y+2}})$ when $d_y=1$.
In a recent paper, \citet{chen2019network} consider multi-product pricing problem with inventory constraints.
Similar to our setting, they assume global concavity and propose an algorithm which also depends on the online learning of gradients and achieves
the regret bound of $\tilde{O}(T^{\frac{4}{5}})$.
The regret is independent of the dimension of the arm space, confirming the insights provided in this paper.
However, the rate of regret bound does not seem to be optimal and the contextual information is not considered.

\textbf{Main results and contributions.}
According to \citet{slivkins2014contextual} and references therein, the optimal regret for the contextual continuum-armed bandits is $\tilde{O}(T^{\frac{d_x+d_y+1}{d_x+d_y+2}})$, when
the payoff function $f(x,y)$ is Lipschitz continuous and $d_x,d_y$ are the dimensions of the context and arm space.
After imposing the structure property that $f(x,y)$ is globally concave in the decision variable $y$ when fixing $x$,
we provide an algorithm that achieves the regret $O(d_y^{\frac{d_x+6}{2(d_x+2)}} T^{\frac{d_x+1}{d_x+2}})$.
Compared to the previous bound, the dimensionality $d_y$ does not increase the regret exponentially when $T$ increases.
The mitigation of the curse of the dimensionality can improve the performance of the algorithm significantly in practice.
For example, in the context-free setting ($d_x=0$), the regret of a ten-dimensional decision variable ($d_y=10$) is $ \tilde{O} (T^{11/12})$ for previous algorithms, while a mere $\tilde{O}(\sqrt{T})$ for our algorithm.
On the other hand, global concavity is a mild assumption,
which is commonly assumed in various applications.
Therefore, the improvement in regret does not come with a substantial sacrifice in the generality of the formulation.

The algorithm is based on binning the contextual space and applying stochastic gradient descent or stochastic approximation in each bin to learn the optimal decision.
Such algorithms are popular in machine learning \citep{bottou2010large,shalev2008svm,shalev2009stochastic, duchi2009efficient}.
In the case of concave functions, the classic algorithms in the learning literature such as UCB or Thompson sampling fail to take into account the special structure and do not seem to be able to achieve the optimal regret.
Instead, gradient-based algorithms, which do not perform well for general functions due to multiple local maxima, can guarantee a surprising dimension-free regret in our setting.
Our results thus convey the message that problem and domain-specific algorithmic design for learning problems may be helpful and beneficial.

{Furthermore, we extend our algorithm to an adaptive binning framework. Such adaptive schemes are also adopted in \citet{slivkins2014contextual, chen2018nonparametric, Perchet2013The}. However, we the first one who design an algorithm combining stochastic approximation with adaptive partition of the context space. Although the new algorithm is much more challenging to analyze, we prove our algorithm achieves the optimal rate. The technical analysis also has important values to other areas, such as simulation optimization, stochastic programming. The adaptive algorithm outperforms the static one in practice, which is supported by the numerical experiment in Section~\ref{numerical}.}




\section{Problem Formulation}
The domain of the unknown payoff function $f(x,y)$ is $x\in \mathcal X\triangleq [0,1]^{d_x}$ and $y\in \mathcal Y\triangleq [0,1]^{d_y}$.
One can interpret $\mathcal X$ and $\mathcal Y$ to be the normalization of some compact sets.
Let $\mathcal{T}=\{1,2,\dots,T\}$ denote the sequence of decision epochs faced by the decision maker.
At the beginning of each epoch $t \in \mathcal{T}$, the contextual information $x_t \in \mathcal{X}$ is revealed to the decision maker.
The contextual information is drawn independently from some unknown distribution\footnote{\citet{slivkins2014contextual} assumes that the context arrivals $x_t$ are fixed before the first round. We follow \citet{Perchet2013The} and assume that $X_t$ are i.i.d.} and therefore denoted by $X_t$.
Then the decision maker chooses an arm $y_t$ in $\mathcal{Y}$.
The payoff in epoch $t$ is a random variable $Z_t$, whose mean is $\mathbb E[Z_t|X_t,y_t]=f(X_t,y_t)$.
We require $Z_t$ to be independent across epochs given $X_t$ and $y_t$.


\textbf{Regret.}
If the payoff function were known, then the optimal decision and average payoff given context $x$ are
\begin{equation*}
    y^{\ast}(x) \triangleq \argmax_{y\in \mathcal Y} f(x,y),\quad f^{\ast}(x) = \max_{y\in\mathcal Y} f(x,y),
\end{equation*}
which is referred to as the oracle.
Since the decision maker does not have access to the unknown function, the total payoff is always lower than that of the oracle in expectation.
A standard performance metric of an algorithm is defined as the regret incurred compared to the oracle.
\begin{equation}
R (T) \triangleq \sum_{t=1}^T \mathbb{E} \left[f^{\ast}(X_t)-f(X_t,y_t)\right].
\end{equation}
Note that the decision made in epoch $t$, $y_t$, is also random, even though the decision maker does not use active randomization.
This is because at each epoch $t$, the decision maker may rely on the information revealed so far to make decisions.
Therefore, $y_t$ may depend on the observed contexts $\left\{X_s\right\}_{s=1}^{t}$, the adopted decisions $\left\{y_s\right\}_{s=1}^{t-1}$ and the realized payoffs $\left\{Z_s\right\}_{s=1}^{t-1}$.
Since $f$ is unknown to the decision maker, the objective is thus to design an algorithm that achieves small regret for a wide class of functions as $T\to\infty$.
One can expect that if $f$ is an arbitrary function such as discontinuous ones, then no algorithm can achieve small regret.
Next, we specify the assumptions that $f$ has to satisfy.


\subsection{Assumptions}\label{sec:assumptions}
We now present a set of assumptions in our setting, which are required to guarantee the existence and good behavior of the gradient estimates.
They are not required by Lipschitz bandits \citep{slivkins2014contextual}.
Most assumptions are mild.

\begin{assumption}[Twice differentiable]\label{asp:2differentiable}
    For all $x \in \mathcal{X}$, the function $f(x,\cdot)$ is twice continuously differentiable on the arm space $\mathcal{Y}$.
\end{assumption}


Besides the existence of gradient,
We assume strong concavity to ensure the global convergence of gradient descent and the uniqueness of the optimal solution.

\begin{assumption}[Strong concavity]\label{asp:concavity}
    There exists a constant $M_1>0$ such that $f(x,y_1) \leq f(x,y_2) + (y_1-y_2)^T \frac{\partial }{\partial y}f(x,y_2) -\frac{1}{2}M_1 \|y_1-y_2\|^2_2 $ for all $x \in \mathcal{X}$ and $y_1,y_2 \in \mathcal{Y}$.
\end{assumption}

An immediate implication of Assumption~\ref{asp:concavity} is the unique optimal solution $y^*(x)=\arg\max_{y \in \mathcal{Y}} f(x,y)$ for any context $x$.
The following assumption makes sure that $y^\ast(x)$ is in the interior of $\mathcal{Y}$, which implies that $\frac{\partial }{\partial y} f(x,y^\ast(x))=0$.

\begin{assumption}[Interior optimal solution]\label{asp:interior}
    For any $x \in \mathcal{X}$, the optimal solution $y^\ast(x)= \arg\max_{y \in \mathcal{Y}} f(x,y)$ satisfies $y^\ast(x) \in int(\mathcal{Y})$.
\end{assumption}
Assumption~\ref{asp:interior} is imposed mainly for technical purposes.
In practice, one may also extend the domain of $y$ to ensure an interior optimal solution.
The next assumption imposes some regularity (H\"{o}lder condition) on the context space.

\begin{assumption}[H\"{o}lder continuity of the context]\label{asp:holder}
    For every $y \in \mathcal{Y}$, the function $f(\cdot,y)$ is H\"{o}lder continuous in $\mathcal{X}$, i.e. $|f(x_1,y)-f(x_2,y)| \leq M_2 \|x_1-x_2\|_2^{\alpha}$ with constant $M_2>0$ and $0<\alpha \leq 1$.
\end{assumption}

H\"{o}lder continuity is a generalization of Lipschitz continuity. It is easy to see that for $\alpha=1$, it is equivalent to Lipschitz continuity.
A Similar condition is also imposed in \citet{Perchet2013The}.
The next assumption makes sure that the random payoff $Z$ behaves normally, which is standard in the literature.
\begin{assumption}[Finite second moment]\label{asp:random-noise}
    For any given $x$ and $y$, the random payoff $Z$ has finite second moment, i.e., there exists a uniform constant $M_3 >0$ such that $\mathbb{E} \left[Z^2 | x,y\right] \leq M_3$.
\end{assumption}
At the beginning of the horizon, the following information is revealed to the decision maker: the domain of the context $\mathcal X$ and the decision variable $\mathcal Y$, the length of the horizon $T$, and the constant $M_1$ defined in the Assumption~\ref{asp:concavity}.

\section{Our Algorithm} \label{sec:our-alg}
There are two components of our algorithm.
To deal with the contexts, we partition the context space into rectangular bins.
When the partition is designed carefully, we are able to conduct context-free learning in each of the bin without significantly increasing the regret.
That is, treat the contexts generated in the same bin equally.
This idea is also adopted by \citet{lu2009showing,rigollet2010nonparametric,Perchet2013The,chen2018nonparametric}.
To find the optimal solution $y^{\ast}(x)$ when the context falls into a particular bin, we use stochastic approximation and the estimated gradient to find the maximum.
Next we elaborate on the two components separately.

\textbf{Binning the context space}.
Discretization and local approximations are probably the most popular method to deal with nonparametric estimation.
Utilizing Assumption~\ref{asp:holder}, one can expect that $f(x_1,\cdot)$ and $f(x_2,\cdot)$ tend to behave similarly, including close maximal values and optimal solutions, when $\|x_1-x_2\|_2$ is small.
Following this intuition, we partition the context space as follows.
We divide each of the $d_x$ dimensions of the context space into $K$ equal intervals.
As a result, the context space $\mathcal X$ divided into $K^{d_x}$ identical hyper-rectangles, referred to as bins.
The partition $\mathcal{B}_K \triangleq \left\{ B_1,\dots,B_{K^{d_x}} \right\}$ is thus a collection of bins of the following form: for $k=(k_1,\dots,k_{d_x}) \in \{1,\dots,K\}^{d_x}$,
\begin{equation}\label{eq:partition}
B_k = \left\{ x \in \mathcal{X}: \frac{k_l-1 }{K} \leq x_l \leq \frac{k_l}{K}, l=1,\dots,d_x \right\}.
\end{equation}
The algorithm thus keeps track of $K^{d_x}$ independent learning sub-problems.
When a context is generated in $B_k$ at some epoch $t$, the exact location of $X_t$ is no longer used as long as the knowledge of $B_k$ is preserved.
The sub-problem $k$ then proceeds with one more epoch while the other sub-problems remain the same.
Therefore, the sub-problem associated with $B_k$ is equivalent to a classic continuum-armed bandit problem without contextual information.

One can clearly see the trade-off in choosing a proper granularity of discretization, represented by $K$.
When $K$ is too small, the algorithm aggregates too much information into a bin and loses accuracy; when $K$ is too large, then there are too many bins and the learning horizon is short for each sub-problem.
Later, we will choose an optimal $K$ to balance the trade-off and obtain small regret.

\textbf{Stochastic Approximation (SA)}.
To solve the sub-problem in a particular bin $B\in \mathcal B_K$, the algorithm treats all contextual information equally as long as a context falls into $B$.
In this case, the oracle for the context-free problem can obtain the average payoff
\begin{equation}\label{eq:fB}
f_B(y)=\mathbb{E} \left[f(X,y) | X \in B\right]
\end{equation}
with the following optimal solution
\begin{equation}
y^\ast(B)=\argmax_{y \in \mathcal{Y}} f_B(y).
\end{equation}

We develops an algorithm based on stochastic approximation (see \citet{Kushner2003Stochastic,chau2015overview} for a comprehensive review) to find $y^{\ast}(B)$.
The basic idea is demonstrated below:
Suppose at epoch $t$ a context $X_{t}$ is generated in bin $B$.
A decision $y_t$ is chosen and a random reward $f(X_t,y_t)$ is observed.
After a number of epochs, another context $X_{t'}$ is observed in $B$ for some $t'>t$.
If the gradient information of $f_B(y)$ at $y_t$ is known, then stochastic approximation could be used to determine $y_{t'}$.
In particular,
\begin{equation*}
    y_{t'} = y_{t} - a_t \nabla f_B(y_{t}).
\end{equation*}
For a properly chosen step size $a_t$, one can show that $y_t$ converges to the optimal solution $y^{\ast}(B)$.

However, there are two pitfalls of when applying SA directly.
First, $y_{t'}$ may be outside the domain $\mathcal Y$.
This issue can be addressed by projecting $y_{t'}$ back to $\mathcal Y$, denoted by the operator $\Pi_{\mathcal Y}$.
Second, the function $f$ is not known to the decision maker, not to mention the gradient $\nabla f_B$.
Thus, we need to estimate the gradient from noisy payoffs $Z_t$.
We use the Kiefer-Wolfowitz (KW) algorithm \citep{Kiefer1952Stochastic} as a subroutine.
After applying $y_t$ and observing $Z_t$, the decision maker explores the neighborhood of $y_t$ and uses a finite-difference method to estimate the gradient.
More precisely, suppose the contexts generated at $t_{d_y}>t_{d_y-1}>\dots>t_1>t$ fall into the same bin $B$.
Our algorithm applies decision $y_t+c e_i$ in epoch $t_i$, where $e_i$ is the unit vector with the $i$th entry equal to 1 and 0 elsewhere.
The step size $c$ will be specified later.
The payoff $Z_{t_i}$ can thus be regarded as an estimator for $f_B(y_t+ce_i)$.
After epoch $t_{d_y}$, the KW algorithm suggests an estimator for $\nabla f_B(y_t)$
\begin{equation}\label{eq:gradient-kw}
\hat{\nabla} f_B(y_t) = \frac{1}{c} \left[Z_{t_1}-Z_t, Z_{t_1}-Z_t,\dots, Z_{t_{d_y}}-Z_t\right]^T
\end{equation}

After $d_y$ more contexts generated in $B$, the algorithm finally moves along the direction of the estimated gradient.
Suppose in epoch $t'>t_{d_y}$, the context $X_{t'}\in B$.
Then the decision $y_{t'}$ is chosen according to
\begin{equation*}
    y_{t'} = \Pi_{\mathcal Y}\left(y_{t} - a_t \hat\nabla f_B(y_{t})\right).
\end{equation*}
Another $d_y$ contexts need to be observed in $B$ in order to estimate $\nabla f_B(y_{t'})$.
The algorithm associated with a single bin is demonstrated in Algorithm~\ref{alg:single-bin}.
To simplify the notation, we only focus on the epochs when the contexts generated are in the same bin and re-order the index by $t=1,2,3,\dots$.
\begin{algorithm}
	\caption{Online KWSA algorithm in one bin}
        \label{alg:single-bin}
	\begin{algorithmic}[]
		\STATE Input: pre-defined step size sequences $a_n,c_n$
		\STATE Initialize: $\tilde{y}_0 \in \mathcal{Y}$
                \STATE $y_1\gets \tilde{y}_0$; observe $Z_1$
                \FOR {$t=2,3,\dots$}
                \STATE $m\gets (t-1) \mod d_y+1$
                \STATE $n\gets (t-1-m)/(d_y+1)$
                \COMMENT{In epoch $t$, the algorithm has estimated the gradient $n$ times}
                \IF{$m\neq 0$}
                \STATE $y_t\gets \tilde{y}_n+c_n e_m$\COMMENT{Use finite difference to estimate the gradient}\label{step:fd}
                \ELSE
		\STATE $G(y_{n-1})\gets \frac{1}{c_{n-1}} \left[ Z_{t-1}-Z_{t-d_y-1}, Z_{t-2}-Z_{t-d_y-1},\dots,Z_{t-d_y}-Z_{t-d_y-1}\right]^T$
		\STATE $\tilde{y}_{n}\gets \Pi_{\mathcal{Y}} \left( \tilde{y}_{n-1}+ a_{n-1} G(y_{n-1})\right)$  \COMMENT{Apply KWSA}
		\STATE $y_t\gets \tilde{y}_n$
                \ENDIF
                \STATE Observe $Z_t$
		\ENDFOR
	\end{algorithmic}
\end{algorithm}

Next we combine the two components for the contextual continuum-armed bandit problem.
From the above description, we keep track of $K^{d_x}$ instances of Algorithm~\ref{alg:single-bin} and updates the number of epochs in each bin separately.
The detailed steps are demonstrated in Algorithm~\ref{alg:KWSA}.
\begin{algorithm}
	\caption{KWSA with binning}
	\label{alg:KWSA}
	\begin{algorithmic}[]
		\STATE Input: $T$, $M_1$
		\STATE Tunable parameters: $K$
		\COMMENT {$K$ is the number of bins}
		\STATE Partition the context space into $\mathcal{B}_K$ as in \eqref{eq:partition}
		\FOR{$t=1,2,\dots,T$}
		\STATE Observe context $X_t$
		\STATE $B \gets \left\{ B \in \mathcal{B}_K : X_t \in B \right\}$ \COMMENT{Determine the bin which $X_t$ belongs to}
		\STATE Apply Algorithm~\ref{alg:single-bin} associated with bin $B$
                \ENDFOR
	\end{algorithmic}
\end{algorithm}
\begin{remark}
    Technically speaking, the finite difference $\tilde{y}_n+c_n e_m$ in Step~\ref{step:fd} of Algorithm~\ref{alg:single-bin} may be outside the domain $\mathcal Y$ and the algorithm needs to adjust for that.
    In that case, we let $y_t \rightarrow \tilde{y}_n-c_n e_m$ which must be inside the domain $\mathcal Y$. And then replace the corresponding difference $Z_{t-i}-Z_{t-d_y-i}$ in $G(y_{n-1})$ by its opposite. After that, all the following analysis remains the same.
\end{remark}

\section{Analysis of the Regret} \label{Ana-reg}
In this section, we provide a roadmap of the analysis.
We first bound the regret incurred in a single bin.
\begin{proposition}\label{prop:regret-single-bin}
    Let $a_n=a n^{-1}$ and $c_n=\delta n^{-\frac{1}{4}}$, where $1/(4M_1)<a<1/(2M_1)$ and $\delta>0$.
    Then the regret of Algorithm 1 in bin $B$ satisfies
    \begin{equation}\label{eq:proposition1}
        \mathbb{E} \left[f_B(y^\ast(B))-f_B(y_t)\right] \leq \frac{\sqrt{d_y }Q(d_y)}{\sqrt{t}}
    \end{equation}
    where $Q$ is a linear function of $d_y$ whose coefficients are independent of $t$. Specifically,
    \begin{equation*}
    Q(d_y)=M_5 \delta^2 + M_5 \max \left\{\mathbb{E} \left[\|\tilde{y}_0-y^*(B)\|_2^2\right], \left(\dfrac{2 \delta M_5 + \sqrt{4 \delta^2 M_5^2+8 d_y a^2 M_3^2 (4a M_1 -1)/\delta^2}}{4 \delta M_5 -1 }\right)^2\right\}
    \end{equation*}
\end{proposition}
Proposition~\ref{prop:regret-single-bin} uses the standard convergence results of KWSA.
However, we need to analyze the property of $f_B(y)$ in \eqref{eq:fB} carefully.
In particular, the assumptions imposed in Section~\ref{sec:assumptions} are for the function $f(x,y)$.
First, we translate the assumptions in Section~\ref{sec:assumptions} of $f(x,y)$ to obtain other crucial properties, including Lipschitz continuity, Lipschitz-continuous gradient and negative-definite Hessian matrix.
Second, we prove the interchangeability of expectation and differentiation of $f(x,y)$ to make sure the properties are extended to $f_B(y)$.
Third, with the regulairty conditions of $f_B(y)$, we apply the asymptotic analysis in the stochastic approximation literature to derive finite time bound of Algorithm~\ref{alg:single-bin}.
More precisely, the left-hand side of \eqref{eq:proposition1} can be bounded by $\|y^\ast(B)-y_t\|_2^2$, because of bounded eigenvalues of the Hessian matrix.
Then we bound $\|y^\ast(B)-y_t\|_2^2$ by a decreasing sequence with convergence rate $t^{-\frac{1}{2}}$.

\begin{remark}
According to Proposition~\ref{prop:regret-single-bin},
when there is not contextual information, Algorithm~\ref{alg:single-bin} achieves a bound $O(d_y^{3/2}\sqrt{T})$ for continuum-armed bandits.
The problem is studied before in the literature and we compare to their results below.
\citet{cope2009regret} shows a similar regret bound $O(\sqrt{T})$ asymptotically.
{Our work has two improvements over his. First, we relax two critical assumptions in his paper, i.e., "the payoff function should be three times continuously differentiable" and "the set of optimal arms $\{y^*(x),x \in \mathcal{X}\}$ should be convex and contain an open ball in $R^{d_y}$". Second, our regret bound is a finite-time bound, while his is asymptotic.}
\citet{bubeck2011x} find that if the smoothness parameter of the payoff functions around the maxima were known, then the near-optimal regret $\tilde{O}(\sqrt{T})$ could be achieved, independent of the dimension of the arm space.
We do not require a certain degree of smoothness for the payoff function; instead, global convexity/concavity is imposed.
In practice, knowing whether the unknown payoff function is convex seems to be more reasonable than knowing how smooth the function is.
In a similar setting, \citet{agarwal2013stochastic} propose an algorithm whose high-probability regret bound is $\tilde{O} (poly(d) \sqrt{T})$.
{Comparing with \citet{agarwal2013stochastic}, we have three advantages. First, we eliminate the logarithmic terms in the bound. Second, we prove the bound with probability 1 instead of $1-1/T$ in their paper. Third, our algorithm is easy to implement.}
\end{remark}

From Proposition~\ref{prop:regret-single-bin}, we know that the regret incurred in one bin is bounded by $O(d_y^{3/2}\sqrt{T})$ if there are $T$ epochs to learn in that bin.
This is because summing up $1/\sqrt{t}$ leads to
\begin{equation*}
    2(\sqrt{T}-1)=\int_1^T \frac{1}{\sqrt{t}}dt \leq \sum_{t=1}^T \frac{1}{\sqrt{t}}dt =1+\sum_{t=2}^T \frac{1}{\sqrt{t}} \leq 1+\int_1^T \frac{1}{\sqrt{t}} dt=2\sqrt{T}-1.
\end{equation*}
To analyze the regret incurred by Algorithm~\ref{alg:KWSA}, we need to aggregate the regret incurred in all the bins in the partition.
Moreover, the optimal solution $y^{\ast}(B)$ for the context-free problem in a bin is still not as good as the oracle $y^{\ast}(X_t)$.
We expect to bound $f(X_t,y^{\ast}(X_t))-f(X_t,y^{\ast}(B))$ by the size the bin and the continuity of $f(x,y)$.
We choose $K=O(T^{\frac{1}{d_x+2\alpha}})$ in Algorithm~\ref{alg:KWSA}.

\begin{theorem}\label{thm:regret}
    For any function $f$ satisfying the assumptions in Section~\ref{sec:assumptions}, the regret by Algorithm~\ref{alg:KWSA} is bounded by
\begin{equation}\label{eq:theorem}
R(T) \leq C d_y^{\frac{\alpha(d_x+6)}{2(d_x+2\alpha)}}T^{\frac{d_x+\alpha}{d_x+2\alpha}}
\end{equation}
for a constant $C$ that is independent of $d_x,d_y,T$.
\end{theorem}
For the most common case of Lipschitz functions, we let $\alpha=1$ and the regret bound becomes
\begin{equation*}
    \tilde{O}(d_y^{\frac{(d_x+6)}{2(d_x+2)}}T^{\frac{d_x+1}{d_x+2}}).
\end{equation*}
It recovers the regret bound in Lipschitz bandit \citep{slivkins2014contextual} with $d_y=0$.
Also notice the fact that when $d_x\ge 2$, $d_y^{\frac{(d_x+6)}{2(d_x+2)}}\le d_y$.
Therefore, no matter how large the dimension of the decision variable $d_y$ is, the dependence of the regret on $d_y$ is at most linear.
Compared to the exponential dependence (i.e., $\tilde{O}(T^{\frac{d_x+d_y+1}{d_x+d_y+2}})$) in the literature, the mild dependence makes our algorithm more suitable for problems with
high-dimensional decision variables.
The significantly improved regret comes at the cost of a more restrictive form of $f(x,y)$, in particular, it has to be globally concave in $y$.
The additional assumption is still reasonable in various applications.
Moreover, the algorithm eliminates the logarithmic terms commonly seen in the literature.


The main steps of the proof are described below. First, the regret incurred in epoch $t$, $\mathbb{E} \left[f^*(X_t)-f(X_t,y_t)\right]$ is bounded by a constant multiplied by the mean square error $\mathbb{E} \left[ \|y_t- y^*(X_t)\|_2^2 \right]$, because of the global convexity.
The distance between $y_t$ and $y^*(X_t)$ is further bounded by $\|y_t-y^*(B)\|_2$ and $\|y^*(B)-y^*(X_t)\|_2$, where the bound of the first term is implied by Proposition~\ref{prop:regret-single-bin}.
For the second term, the discretization error incurred by binning, is bounded by the diameter of bin $B$.
Second, after deriving the regret incurred in one epoch, the bound of total regret could be obtained by summing up the regret in all bins.
The worst case in terms of the regret is when the covariates are generated evenly in the bins, and the best case is when the covariates are generated in a single bin.
Therefore, suppose each of the $K^{d_x}$ bins observes $T/K^{d_x}$ covariates and we can compute the aggregate regret for this worst case.
Finally, we need to minimize the regret over the number of bins $K$.
The tunable parameter $K$ can be regarded as a balance between exploration and exploitation. When the bin is large, there is enough observations in the bin that the selected arm is close to its optimum. However, due to the large diameter of the bin, its optimum may be far away from the optimum respect to one specific covariate in the bin. On the other hand, when the bin is small, the distance between the optimum of the bin and optimum of one
covariate is quite small, but the arms chosen by Algorithm~\ref{alg:single-bin}, may be far from the bin’s optimum. To balance the trade-off and obtain smallest regret, the number of bins is chosen to be $K=O(T^{\frac{1}{d_x+2\alpha}})$.
\begin{remark}
    Another type of stochastic approximation is referred to as the Robbins-Monro \citep{robbins1951stochastic} algorithm.
    Different from KW, RM algorithm requires an oracle to return the unbiased estimators for the gradient of $f_B(y)$ for any given $y$.
    The unbiased estimator can be used to replace $\hat{\nabla} f_B(y_{t-1})$ in \eqref{eq:gradient-kw}.
    As a result, the convergence rate of RM is better than KW and the regret bound in Theorem~\ref{thm:regret} can be further improved.
    However, the information of an unbiased estimator for the gradient is a somewhat unrealistic scenario, and thus we do not present the regret for the RM algorithm in this paper.
\end{remark}

\section{Extension to Adaptive Binning}

There is one potential pitfall in the binning algorithm proposed in Section~\ref{sec:our-alg}. The size of the bins is predefined before the algorithm starts. So the bias due to binning will not decrease as more context observed in the bin. To remedy the pitfall, we propose an adaptive binning algorithm which reduces the binning bias by sequentially splitting the bin into smaller ones. More specifically, the adaptive algorithm has three advantages over the static one. First, taking advantage of sequentially arrived context, it partitions the space adaptively to the distribution of context to maintain a finer partition in popular regions of context. Second, when binning adaptively, the observations in a parent bin provide some information for all offspring bins. Such a pooling effect makes the exploration more effective. While in the setting of static binning, the information learned from observations is only restricted in its own bin. Third, as mentioned in \cite{slivkins2014contextual}, the adaptive algorithm can be applied to solve other MAB problem, such as MAB with stochastically evolving payoffs and sleeping bandits.

\textbf{Adaptive binning the context space}. 
Rather than fixing all bins at the beginning, the adaptive algorithm splits bins sequentially. When enough covariates are observed in the bin, we bisect it in all the $d_x$ dimensions to obtain $2^{d_x}$ child bins. Specifically, for a bin $B$ with boundaries $a_i$ and $b_i$ for $i=1,\ldots,d_x$, its children are indexed by $i \in \{0,1\}^{d_x}$ and have the form 
\begin{equation*}
B_i=\left\{ x: a_j \leq x_j < \frac{a_j+b_j}{2} \ \rm{if} \ i_j=0, \frac{a_j+b_j}{2} \leq x_j < b_j \ \rm{if} \ i_j=1, \ j=1,\ldots,d \right\}
\end{equation*}
Our algorithm starts with a root bin $B_0$, whose depth level is denoted by 0. When covariates are observed $n_0$ times in $B_0$, it is split into $2^{d_x}$ level 1 bins. For such a bin $B_1$, it will be split into $2^{d_x}$ level 2 bins once covariates are observed further $n_1$ times in the bin. So on so forth until the bin reaches the deepest depth $L$. When a bin is at level $L$, the algorithm no longer splits it and simply applies the decision of KW (Kiefer-Wolfowitz) algorithm whenever a covariate is generated in it.

There are two critical hyper-parameters required to be carefully designed before the beginning. First, $n_l$---the maximal number of observations collected in a level-$l$ bin before partition---balances the trade-off between bias due to binning and learning horizon in each bin. When $n_l$ is too small, the learning horizon for bins in level k is too short that the solution obtained by KW algorithm may not be close to the optimum. When $n_l$ is too large, the bias is too large which makes the KW algorithm lose accuracy. Second, $L$---the maximal level of bins---balances the trade-off between adding more child bins and the time horizons in the deepest bins. When a bin in the deepest level is split, the bias of the bin becomes smaller, but the total number of the bins increases by $2^{d_x}$.

A crucial step in the algorithm is to determine what information to inherit when a bin is split into child bins. Since the last decision in parent bin is the best indicator for its optimum, we adopt it as the initial decision in child bins. 

Next, we combine the adaptive binning algorithm with Algorithm~\ref{alg:single-bin} mentioned in section \ref{sec:our-alg}. The detailed steps are demonstrated in Algorithm~\ref{alg:KWSA-adap}.

\begin{algorithm}
	\caption{KWSA with adaptive binning}
	\label{alg:KWSA-adap}
	\begin{algorithmic}[]
		\STATE Input: $T$, $M_1$
		\STATE Tunable parameters: $L$, $n_l$ for $l=0,\cdots,L$
		\STATE Initialize: partition $\mathcal{B} \leftarrow B_{\phi}$ 
		\COMMENT {Initialize the partition set with root bin}
		\FOR{$t=1,2,\dots,T$}
		\STATE Observe context $X_t$
		\STATE $B \leftarrow \left\{ B \in \mathcal{B} : X_t \in B \right\}$ \COMMENT {The bin in the partition that $X_t$ belongs to}
		\STATE $l \leftarrow l(B), \ N(B) \leftarrow N(B)+1 $
		\COMMENT {Determine the level and update the number of covariates observed in $B$}
		\IF{$l < L$} 
		\IF{$N(B)<n_l$}
		\STATE Apply Algorithm~\ref{alg:single-bin} associated with bin $B$
		\ELSE
		\STATE $y^*_{B} \leftarrow y_{N(B)}$ \COMMENT{Define the empirically-optimal decision in bin $B$ as the last decision}
		\STATE $\mathcal{B} \leftarrow (\mathcal{B} \backslash B) \cup C(B)$ \COMMENT{Update the partition by removing $B$ and adding its children}
		\FOR{$B' \in C(B)$}
		\STATE $N(B') \leftarrow 0$ \COMMENT{Initialization for each child bin}
		\STATE $\tilde{y}_0 \leftarrow y^*_{B}$ \COMMENT{Set initial decision as the last one in parent bin}
		\STATE Initialize Algorithm~\ref{alg:single-bin} associated with bin $B'$
		\ENDFOR	
		\ENDIF
		\ELSE
		\STATE Apply Algorithm~\ref{alg:single-bin} associated with bin $B$
		\ENDIF
		\ENDFOR
	\end{algorithmic}
\end{algorithm}

\textbf{Analysis of the regret}. The adaptive binning algorithm is much more challenging to analyze because more hyper-parameters need to be carefully designed. Under the choice of $L=O(\frac{\log T}{(d+2 \alpha ) \log 2})$ and $n_l=2^{2 \alpha l}$, we prove that the adaptive binning algorithm achieves the same order of regret upper bound as the static one. 

\begin{theorem}\label{thm:regret-adap}
	For any function $f$ satisfying the assumptions in Section~\ref{sec:assumptions}, the regret by Algorithm~\ref{alg:KWSA-adap} is bounded by
	\begin{equation}\label{eq:theorem-adap}
	R(T) \leq C d_y^{\frac{3}{2}+\frac{2 d_x}{3 (d_x+2\alpha)}}T^{\frac{d_x+\alpha}{d_x+2\alpha}}
	\end{equation}
	for a constant $C$ that is independent of $d_x,d_y,T$.
\end{theorem}

Intuitively, the adaptive algorithm should outperform the static one because it refines more partitions in popular regions of context. However, Theorem~\ref{thm:regret-adap} suggests that the adaptive and static algorithms achieve the same asymptotic rate of regret. That's because the performance measure is an upper bound no matter for any distribution of covariates. So the worst-case performance is considered when counting the total regret. The worst-case for both the adaptive and static algorithm is that the covariates are distributed uniformly in the whole space, which exactly eliminates the advantage of adaptive binning. Thus, the benefit of the adaptive algorithm is not reflected in the asymptotic rate of regret. However, the adaptive algorithm performs better than the static one in practice (see Section~\ref{numerical}).

{Theoretical analysis is much more difficult than the static one}
\begin{remark}
	As mentioned in Section~\ref{Ana-reg}, some prior work also attain the optimal regret rate in context-free setting. But their algorithms or analysis are difficult to extend to an adaptive scheme. Specifically,  \citet{cope2009regret} assumes the set of optimal arms $\{y^*(x),x \in \mathcal{X}\}$ should be convex and contain an open ball in $R^{d_y}$. This assumption may be violated easily when the context is restricted to a sub-area.  \citet{agarwal2013stochastic} propose an algorithm obtaining the optimal rate with a probability $1-1/T$. When combining with static binning, their good-performance probability $1-T^{\frac{2}{d_x+2}}$ decreases with $d_x$. It's more challenging to make sure that their algorithm still attains the optimal bound with high-probability when extending to the adaptive one.
\end{remark}

\section{Numerical Experiment} \label{numerical}
In this section, we apply various algorithms to a numerical example with $d_y=2$ and $d_x=1$. In the context space $[0,1]$, there are two community centers: $0.1, 0.9$, corresponding to the two payoff functions $f_1(y_1,y_2)=-(y_1-y_2)^2-(y_1-0.5)^2$ and $f_2(y_1,y_2)=-(y_1-2 y_2)^2-(y_2-1/3)^2$. For a context $x$, its distance to the two centers are denoted $d_1=|x-0.1|$ and $d_2=|x-0.9|$. The total payoffs are defined as
\begin{equation}
f(x,y_1,y_2)=\frac{1}{1/d_1+1/d_2} \left( \frac{1}{d_1} f_1(y_1,y_2)+ \frac{1}{d_2} f_2(y_1,y_2)\right)
\end{equation}
which could be checked satisfying the assumptions in Section ~\ref{sec:assumptions} (details in supplementary material).

\begin{figure}[!htp]
	\centering
	\includegraphics[width=8cm]{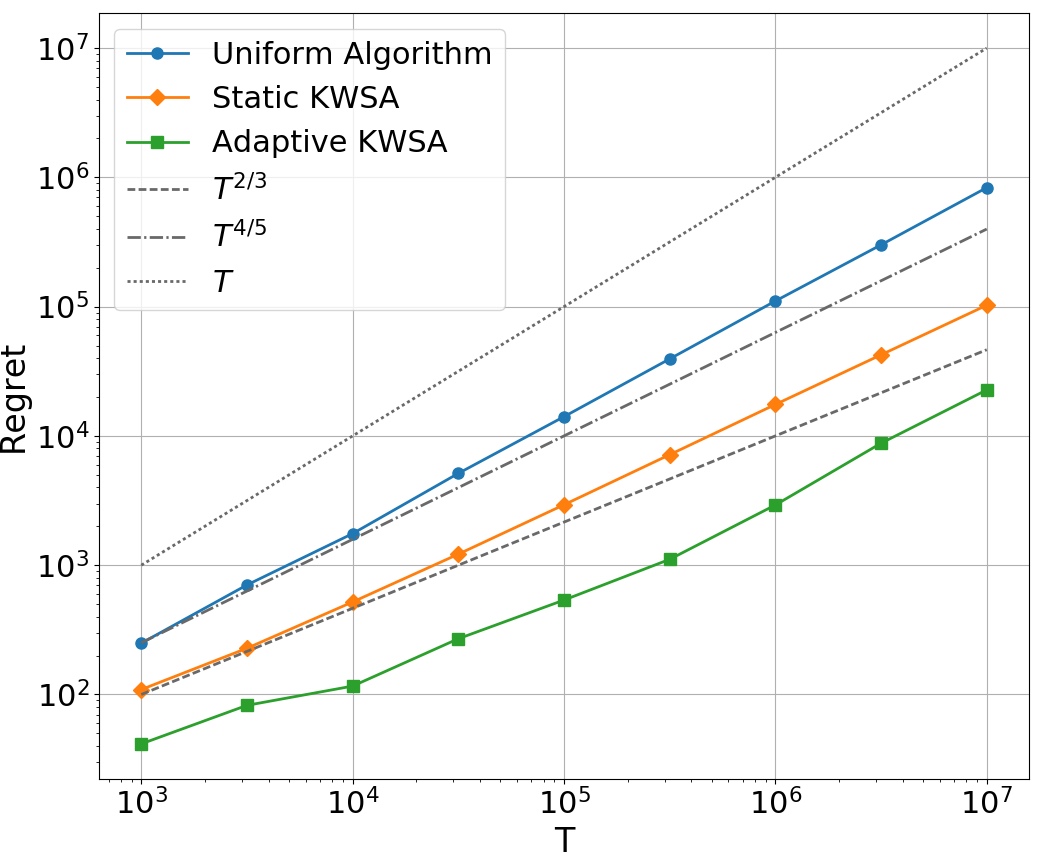}\\
	\caption{\small{Regret against T in a log-log plot}}
	\label{fig:numerical}
\end{figure}

We compare our algorithms (KWSA with static and adaptive binning) with the "uniform algorithm" \citep{slivkins2014contextual}, which simply discretizes the covariate and arms space then apply UCB to sub-problems. With correctly tuned parameters, the "uniform algorithm" attains the regret bound $O(T^{\frac{d_x+d_y+\alpha}{d_x+d_y+2\alpha}})$  ($O(T^{4/5})$ in the example). While our algorithms attain the regret bound $O(T^{2/3})$. The result shows in a log-log plot in Figure~\ref{fig:numerical}. We see that the "uniform algorithm" has a similar growth rate to $T^{4/5}$ and our algorithms have a similar growth rate to $T^{2/3}$. So our algorithms perform much better than "uniform algorithm". Besides that, the adaptive algorithm shows significant improvement over the static one.

\begin{remark}
Comparing to the existing work \citep{slivkins2014contextual} using the adaptive partition scheme, our algorithms are more convenient to implement and show low computational complexity. In the numerical example, when total time horizon $T=10^7$, the computing times for "uniform algorithm", Static KWSA, Adaptive KWSA are 12mins, 6mins and 7mins respectively, using a PC with 3.4GHz Intel Core i7 CPU and 8GB memory. That implies our algorithms reduce the computational burden even comparing to the simplest one. And using adaptive scheme only slightly increases the computing time.
\end{remark}

\section{Conclusion and Future Work}

In this paper, we study the continuum-armed bandit problem under contextual information, where the context space and arms space are both continuous. After assuming the curvature of payoff functions, strong convexity and second-order smoothness, we propose a novel algorithm combining stochastic approximation with binning partition framework and obtain a much better regret than existing literature. Surprisingly, our method achieves a dimension-free result. {Moreover, we generalize our algorithm to an adaptive one that maintains the asymptotic regret bound but performs much better in practice.}   

In the future, we will investigate how to reduce the effect incurred by context space. In other words, whether the curvature conditions corresponding to  covariates can be utilized to improve the regret bound. If so, another nonparametric estimation framework other than binning partition is required to solve the curse of dimensionality problem. It's an open problem that how to  incorporate nonparametric statistic learning methods in reducing the growth rate of the regret with respect to the dimension of context space.

\bibliographystyle{abbrvnat} 
\bibliography{myrefs} 

\section{Appendix}

\subsection{Proof of Proposition 1}
\begin{proposition}
	Let $a_n=a n^{-1}$ and $c_n=\delta n^{-\frac{1}{4}}$, where $1/(4M_1)<a<1/(2M_1)$ and $\delta>0$.
	Then the regret of Algorithm 1 in bin $B$ satisfies
	\begin{equation}
	\mathbb{E} \left[f_B(y^\ast(B))-f_B(y_t)\right] \leq \frac{\sqrt{d_y }Q(d_y)}{\sqrt{t}}
	\end{equation}
	where $Q$ is a linear function of $d_y$ whose coefficients are independent of $t$. Specifically,
	\begin{equation*}
	Q(d_y)=M_5 \delta^2 + M_5 \max \left\{\mathbb{E} \left[\|\tilde{y}_0-y^*(B)\|_2^2\right], \left(\dfrac{2 \delta M_5 + \sqrt{4 \delta^2 M_5^2+8 d_y a^2 M_3^2 (4a M_1 -1)/\delta^2}}{4 \delta M_5 -1 }\right)^2\right\}
	\end{equation*}
\end{proposition}

At a high level the proof first extends assumptions in section 2 to more properties of mean payoff function $f$ in Lemma~\ref{Lemma:1}. Then we prove theses assumptions and properties are also satisfied by the payoff function $f_B$ in Lemma~\ref{Lemma:3}. Finally, Proposition 1 can be proved using these properties. 

\begin{lemma}[Properties of $f(x,y)$] \label{Lemma:1} According to assumptions 1-3, we have that,
	\begin{description}
		\item[(1)] $f(x,y)$ is Lipschitz continuous in $y$ with a constant $M_4$, i.e. $|f(x,y_1)-f(x,y_2)| \leq M_4 \|y_1-y_2\|_2$.
		\item[(2)] $\frac{\partial }{\partial y} f(x,y)$ is Lipschitz continuous in $y$ with a constant $M_5$, i.e. $\|\frac{\partial }{\partial y} f(x,y_1)- \frac{\partial }{\partial y} f(x,y_2)\| \leq M_5 \|y_1-y_2\|_2$.
		\item[(3)] For any $x \in \mathcal{X}$ and $y \in \mathcal{Y}$, the Hessian matrix $\frac{\partial^2 }{\partial y^2} f(x,y)$ is negative definite and all the eigenvalues are in the interval $[-M_5,-M_1]$, i.e. $-M_5I \preceq \frac{\partial^2 }{\partial y^2} f(x,y) \preceq -M_1I$.
		\item[(4)] For every $x \in \mathcal{X}$, the function $f(x,y)$ has a unique maximizer $y^*(x) \in int(\mathcal{Y})$, i.e. there exists a unique $y^*(x) = \argmax_{y \in \mathcal{Y}} f(x,y)$ and $\frac{\partial }{\partial y} f(x,y^*(x))=0$.
	\end{description}
\end{lemma}

\textbf{Proof of Lemma 1.}
\begin{description}
	\item[(1)] Since $\frac{\partial }{\partial y} f(x,\cdot)$ is a continuous function on a convex set $\mathcal{Y}$, there exists a constant $M_1$ s.t $\|\frac{\partial }{\partial y} f(x,\cdot)\|_2 \leq M_1$. Then by generalized mean value theorem (Theorem 9.19 in \cite{rudin1964principles}), $|f(x,y_1)-f(x,y_2)| \leq M_1 \|y_1-y_2\|_2$ for all $x \in \mathcal{X}$ and $y_1,y_2 \in \mathcal{Y}$.
	\item[(2)] Since $\frac{\partial^2 }{\partial y^2} f(x,\cdot)$ is a continuous function on a convex set $\mathcal{Y}$, there exists a constant $M_5$ s.t $\|\frac{\partial^2 }{\partial y^2} f(x,\cdot)\|_2 \leq M_5$. Then by generalized mean value theorem, $\|\frac{\partial }{\partial y} f(x,y_1)- \frac{\partial }{\partial y} f(x,y_2)\|_2 \leq M_5 \|y_1-y_2\|_2$ for all $x \in \mathcal{X}$ and $y_1,y_2 \in \mathcal{Y}$.
	\item[(3)] 	Notice that $f(x,y)$ is a strongly concave function, for every $x \in \mathcal{X}$ and $y_1,y_2 \in \mathcal{Y}$,
	\begin{equation*}
	f(x,y_1) \leq f(x,y_2) + (y_1-y_2)^T \frac{\partial }{\partial y} f(x,y_2)-\frac{1}{2}M_1 \|y_1-y_2\|^2_2 
	\end{equation*}
	By second order Taylor expansion, there exists a $y_0= \beta y_1+(1-\beta) y_2, \ \beta \in [0,1]$ such that,
	\begin{equation*}
	f(x,y_1) = f(x,y_2) + (y_1-y_2)^T \frac{\partial }{\partial y} f(x,y_0)+\frac{1}{2} (y_1-y_2)^T \frac{\partial^2 }{\partial y^2} f(x,y_0) (y_1-y_2)
	\end{equation*}
	Thus,
	\begin{equation*}
	\frac{1}{2} (y_1-y_2)^T \frac{\partial^2 }{\partial y^2} f(x,y_0) (y_1-y_2) \leq -\frac{1}{2} M_1 \|y_1-y_2\|^2_2
	\end{equation*}
	and 
	\begin{equation*}
	\frac{\partial^2 }{\partial y^2} f(x,y_0) \preceq -M_1 I
	\end{equation*}
	Again by twice continuous differentiability of $f_y(x,\cdot)$, we have
	\begin{equation*}
	\frac{\partial^2 }{\partial y^2} f(x,y_0) \rightarrow \frac{\partial^2 }{\partial y^2} f(x,y) \ as \ y_0 \rightarrow y
	\end{equation*}
	So for all $y \in \mathcal{Y}$,
	\begin{equation*}
	\frac{\partial^2 }{\partial y^2} f(x,y) \preceq -M_1 I
	\end{equation*}
	Recall that $\frac{\partial^2 }{\partial y^2} f(x,\cdot)$ is a continuous function on a convex set $\mathcal{Y}$ and $\|\frac{\partial^2 }{\partial y^2} f(x,\cdot)\|_2 \leq M_5$.
	Then for any $y \in \mathcal{Y}$,
	\begin{equation*}
	-M_5I \preceq \frac{d^2 }{d y^2} f_B(y) \preceq - M_1 I, \ M_1 \leq \|\frac{d^2 }{d y^2} f_B(y)\|_2 \leq M_5
	\end{equation*}
	\item[(4)] Assuming there exists $y_1 \neq y_2$ satisfying 
	\begin{equation*}
	f(x,y_1)=f(x,y_2)=\max_{y \in \mathcal{Y}} f(x,y)
	\end{equation*}
	By the strong concavity of $f(x,y)$,
	\begin{equation*}
	2f(x,\frac{y_1+y_2}{2}) \geq f(x,y_1)+f(x,y_2)=2 \max_{y \in \mathcal{Y}} f(x,y)
	\end{equation*}
	which contradicts with the definition of $y^*(x)$. Thus there only exists one maximizer $y^*(x)$.
	
	Next, if $\frac{\partial }{\partial y} f(x,y^*(x)) \neq 0$, we can find a small step $l$ such that $y'=y^*(x)+l \frac{\partial }{\partial y} f(x,y^*(x)) \in \mathcal{Y}$. Then $f(x,y') > f(x,y^*(x))$ which contradicts with the definition of $y^*(x)$. So $\frac{\partial }{\partial y} f(x,y^*(x))=0$.
\end{description}

Before proving the smoothness and convexity conditions of $f_B$, we first provide condition for interchangeability of expectation and derivative in Lemma~\ref{Lemma:2}.

\begin{lemma}[Pathwise Method]\label{Lemma:2} Assume for every $x$, $f(x,\cdot)$ is differentiable on $y \in \mathcal{Y}$ and Lipschitz continuous with constant $M_1$. Then $\frac{\partial }{\partial y} \mathbb{E}_X [f(X,y)]=\mathbb{E}_X [\frac{\partial }{\partial y} f(X,y)]$.
\end{lemma}

\textbf{Proof of Lemma 2.}

Considering one certain dimension $y_i$, if we can prove $\frac{\partial}{\partial y_i} \mathbb{E}_X [f(X,y)]=\mathbb{E}_X [\frac{\partial}{\partial y_i} f(X,y)]$, then Lemma 2 is obvious.
\begin{equation*}
\begin{split}
\frac{\partial}{\partial y_i} \mathbb{E}_X [f(X,y)]
& = \lim_{h \rightarrow 0} \dfrac{\mathbb{E}_X[f(X,y+h e_i)]-\mathbb{E}_X[f(X,y)]}{h}\\
& = \lim_{h \rightarrow 0} \mathbb{E}_X \left[ \dfrac{f(X,y+h e_i)-f(X,y)}{h}\right]\\
& = \mathbb{E}_X \left[ \lim_{h \rightarrow 0} \dfrac{f(X,y+h e_i)-f(X,y)}{h}\right] \\
& = \mathbb{E}_X \left[ \frac{\partial}{\partial y_i} f(X,y)\right]\\
\end{split}
\end{equation*}
In the first equality, $e_i$ denotes the unit vector with the i-th entry 1. The third equality is supported by dominated convergence theorem, where $\dfrac{|f(X,y+h e_i)-f(X,y)|}{h} \leq M_5$.

\begin{lemma} [Optimal arm in hypercube]\label{Lemma:3}
	For any hypercube $B \subset \mathcal{X}$, including a singleton $B=\{x\}$, define the function $f_B(y) \triangleq \mathbb{E} \left[f(X,y)|X \in B\right]$. According to assumptions 1-3 ,we have that for any $B$,
	\begin{description} 
		\item[(1)] $f_B(y)$ is twice continuously differentiable on the convex set $y \in \mathcal{Y}$. Additionally, expectation and gradient are exchangeable, i.e. $\nabla f_B(y) =\mathbb{E} [\frac{\partial }{\partial y} f(X,y)|X \in B] $ and $\frac{d^2 }{d y^2} f_B(y)  = \mathbb{E} \left[\frac{\partial^2 }{\partial y^2} f(X,y)|X \in B\right]$.
		\item[(2)] $f_B(y)$ is strongly concave in $y$ with the same constant of $f(x,y)$, i.e. $f_B(y_1) \leq f_B(y_2) + (y_1-y_2)^T \nabla f_B(y_2)-\frac{1}{2} M_1 \|y_1-y_2\|^2_2 $ for all $y_1,y_2 \in \mathcal{Y}$
		\item[(3)] $f_B(y)$ maintains Lipschitz continuous property of $f(x,y)$ with the same constant, i.e. $|f_B(y_1)-f_B(y_2)| \leq M_4 \|y_1-y_2\|_2$ for all $y_1,y_2 \in \mathcal{Y}$
		\item[(4)] $f_B(y)$ maintains Lipschitz gradient property of $f(x,y)$ with the same constant, i.e. $\|\nabla f_B(y_1)-\nabla f_B(y_2)\|_2 \leq M_5 \|y_1-y_2\|_2$ for all $y_1,y_2 \in \mathcal{Y}$.
		\item[(5)] For all $y \in \mathcal{Y}$, the Hessian matrix of $f_B(y)$ is negative definite and all the eigenvalues are in the interval $[-M_5,-M_1]$, i.e. $-M_5I \preceq \frac{d^2 }{d y^2} f_B(y) \preceq -M_1 I$. 
		\item[(6)] The function $f_B(y)$ has a unique maximizer $y^*(B) \in int (\mathcal{Y})$, i.e. there exists a unique $y^*(B)=\argmax_{y \in int(\mathcal{Y})} \mathbb{E} \left[f(X,y)|X \in B\right]$, and $\nabla f_B(y^*(B))=0$.   
	\end{description}
\end{lemma}

\textbf{Proof of Lemma 3.} 
\begin{description}
	\item[(1)] 
	Since  $f(\cdot,y)$ are differentiable and $f(\cdot,y)$ are Lipschitz continuous with constant $M_1$, according to pathwise method, $\nabla f_B(y)=\frac{\partial }{\partial y} \mathbb{E} [f(X,y)|X \in B] =\mathbb{E} [\frac{\partial }{\partial y} f(X,y)|X \in B] $. $\nabla f_B(y)$ exists for the reason that $\frac{\partial }{\partial y} f(x,y)$ exists for any $x$.
	Similarly, as $\frac{\partial }{\partial y} f(\cdot,y)$ are differentiable and Lipschitz continuous with constant $M_5$, we have 
	\begin{equation*}
	\begin{split}
	\frac{d^2 }{d y^2} f_B(y)
	&= \frac{\partial^2 }{\partial y^2} \mathbb{E} \left[f(X,y)|X \in B\right]\\
	&= \frac{\partial }{\partial y} \mathbb{E} \left[\frac{\partial }{\partial y} f(X,y)|X \in B\right]\\
	&= \mathbb{E} \left[\frac{\partial^2 }{\partial y^2} f(X,y)|X \in B\right]\\
	\end{split}
	\end{equation*}
	The continuity of $\frac{d^2 }{d y^2} f_B(y)$ is the consequence of $\frac{\partial^2 }{\partial y^2} f(x,y)$ continuous in $y$.
	\item[(2)]
	Strong concavity is obvious because expectation operator maintains linear relationship.
	\begin{equation*}
	\begin{split}
	f_B(y_1)
	&= \mathbb{E} [f(X,y_1) |X \in B]\\
	& \leq \mathbb{E} \left[f(X,y_2)+(y_1-y_2)^T \frac{\partial }{\partial y} f(X,y_2) -\frac{1}{2} M_1 \|y_1-y_2\|_2^2 \Big| X \in B \right]\\
	& = f_B(y_2) +(y_1-y_2)^T \nabla f_B(y_2) -\frac{1}{2} M_1 \|y_1-y_2\|_2^2\\
	\end{split}
	\end{equation*}
	\item[(3)]
	\begin{equation*}
	\begin{split}
	|f_B(y_1)-f_B(y_2)|
	&= \left|\mathbb{E} \left[f(X,y_1)|X \in B\right] - \mathbb{E} \left[f(X,y_2)|X \in B\right] \right| \leq \mathbb{E} \left[|f(X,y_1)-f(X,y_2)| \Big| X \in B\right] \\
	& \leq \mathbb{E} \left[ M_4 \|y_1-y_2\|_2 \right]= M_4 \|y_1-y_2\|_2
	\end{split}
	\end{equation*}
	\item[(4)] 
	\begin{equation*}
	\begin{split}
	\|\nabla f_B(y_1)-\nabla f_B(y_2)\|_2
	&= \left\| \mathbb{E} \left[\frac{\partial }{\partial y} f(X,y_1)|X \in B \right] -\mathbb{E} \left[\frac{\partial }{\partial y} f(X,y_2)|X \in B \right]\right\|_2\\
	& \leq \mathbb{E} \left[ \left\|\frac{\partial }{\partial y} f(X,y_1)-\frac{\partial }{\partial y} f(X,y_2)\right\|_2 | X \in B \right] \\
	& \leq \mathbb{E} \left[ M_5\|y_1-y_2\|_2\right]\\
	& \leq M_5 \|y_1-y_2\|_2\\
	\end{split}
	\end{equation*}
	\item[(5)] 
	Since $f_B(y)$ has the twice continuous differentiable and strong concave property of $f(x,y)$, this property can be proved following the same way of Lemma~\ref{Lemma:1}(3).
	\item[(6)] Assuming there exist $y_1 \neq y_2$ satisfying
	\begin{equation*}
	f_B(y_1)=f_B(y_2)=\max_{y \in \mathcal{Y}} \mathbb{E} \left[f(X,y) | X \in B\right]
	\end{equation*} 
	By the strong concavity of $f(x,y)$,
	\begin{equation*}
	2 f(x,\frac{y_1+y_2}{2}) > f(x,y_1)+f(x,y_2)
	\end{equation*} 
	Thus,
	\begin{equation*}
	\begin{split}
	2 f_B \left(\frac{y_1+y_2}{2} \right)
	& =2\mathbb{E} \left[f(X,\frac{y_1+y_2}{2}) | X \in B\right] \\
	& > \mathbb{E} \left[f(X,y_1) | X \in B\right]+\mathbb{E} \left[f(X,y_2) | X \in B\right]=2 f_B(y^*(B))\\
	\end{split}
	\end{equation*}
	which contradicts with the definition of $y^*(B)$. Thus there is only one maximizer $y^*(B)$.
	
	To prove the second part $\nabla f_B(y^*(B))=0$, we define that $y^*(B)=\left[ y_1^*(B), \cdots, y_{d_y}^*(B) \right]^T$ and $y^*(x)=\left[ y_1^*(x), \cdots, y_{d_y}^*(x) \right]^T$. Recall that $\mathcal{Y}$ is a convex set $[0,1]^{d_y}$, the lower bounds and upper bounds in each dimension are denoted by $[y_1^l,y_1^h], \cdots, [y_{d_y}^l,y_{d_y}^h]$. In each dimension $i$, we define the minimum distance between $y_i^*(x)$ and $y_i^l$ as $\delta_i^l = \inf_{x \in B} \Big| y_i^*(x)-y_i^l \Big|$. Similarly,  $\delta_i^h = \inf_{x \in B} \Big| y_i^*(x)-y_i^h \Big|$. $y^*(x) \in int(\mathcal{Y})$, then $\delta_i^l \geq 0$ and $\delta_i^h \geq 0$. Considering that $f(x,y)$ is a strongly concave function in $y$, $\dfrac{\partial f(x,y)}{\partial y_i} \geq 0$ for $y_i \in \left[y_i^l, y_i^l+ \delta_i^l \right]$ and $x \in B$ and $\dfrac{\partial f(x,y)}{\partial y_i} \leq 0$ for $y_i \in \left[y_i^h- \delta_i^h, y_i^h \right]$. So $\dfrac{\partial f_B(y)}{\partial y_i} \geq 0$ for $y_i \in \left[y_i^l, y_i^l+ \delta_i^l \right]$ and $\dfrac{\partial f_B(y)}{\partial y_i} \leq 0$ for $y_i \in \left[y_i^h- \delta_i^h, y_i^h \right]$. Recall that $\dfrac{\partial f_B(y)}{\partial y_i}$ is a continuous function. Then, there exists $y_i^* \in \left[y_i^l, y_i^h\right]$ such that $\dfrac{\partial f_B(y)}{\partial y_i}\Big|_{y_i=y_i^*} =0$.
\end{description}

\begin{lemma}\label{Lemma:4} Suppose the sequence $b_n$ satisfies
	\begin{equation*}
	b_{n+1} \leq \left(1-\frac{\alpha}{n}\right) b_n + \beta n^{-\frac{5}{4}} \sqrt{b_n} + \omega n^{-\frac{3}{2}}
	\end{equation*}
	Let $\lambda=\max\{b_1,\lambda_0\}$, where 
	\begin{equation*}
	\lambda_0=\left(\dfrac{\beta+\sqrt{\beta^2+2 \omega (2 \alpha-1)}}{2 \alpha -1}\right)^2
	\end{equation*}
	Then, by induction, we have $b_n \leq \lambda n^{-\frac{1}{2}}$.
\end{lemma}

\textbf{Proof of Lemma 4.}

We prove by induction. It is easy to see that it hold for $n=1$. For any $n=1,2,\cdots,$ suppose that $b_n \leq \lambda n^{-\frac{1}{2}}$. Because $1-\alpha/n >0$ due to $\alpha <1$,
\begin{equation*}
\begin{split}
b_{n+1} 
& \leq \left(1-\frac{\alpha}{n}\right) \lambda n^{-\frac{1}{2}}+ \beta \sqrt{\lambda} n^{-\frac{3}{2}}+ \omega n^{-\frac{3}{2}}\\
& = \lambda n^{-\frac{1}{2}} - \left(\alpha \lambda - \beta \sqrt{\lambda} -\omega \right) n^{-\frac{3}{2}}\\
& = \lambda n^{-\frac{1}{2}}-\frac{\lambda}{2} n^{-\frac{3}{2}} -\frac{1}{2} \left[(2\alpha-1) \lambda -2 \beta \sqrt{\lambda} -2 \omega \right] n^{-\frac{3}{2}}\\
& \leq \lambda \left(n^{-\frac{1}{2}}-\frac{1}{2} n^{-\frac{3}{2}}\right)
\end{split}
\end{equation*}
where the last inequality follows form the definition of $\lambda_0$ that satisfies $(2\alpha-1) z- 2 \beta \sqrt{z}-2 \omega \leq 0$ for any $z \geq \lambda_0$. Let $g(x)=x^{-\frac{1}{2}}$. Then, $g'(x)=-\frac{1}{2}x^{-\frac{3}{2}}$. Notice that $g(x)$ is convex. Then,
\begin{equation*}
g(x')-g(x) \geq g'(x) (x'-x)
\end{equation*}
Then,
\begin{equation*}
(n+1)^{-\frac{1}{2}}-n^{-\frac{1}{2}} = g(n+1)-g(n) \geq g'(n) =-\frac{1}{2} n^{-\frac{3}{2}}
\end{equation*}
Therefore,
\begin{equation*}
n^{-\frac{1}{2}}-\frac{1}{2}n^{-\frac{3}{2}} \leq (n+1)^{-\frac{1}{2}}
\end{equation*}
Then, we have $b_{n+1} \leq \lambda (n+1)^{-\frac{1}{2}}$. This concludes the induction proof. $\Box$

\textbf{Proof of Proposition 1.}
According to the dimension of arm space, the time epochs is divided into periods: period 1=$\{1,\dots,d_y\}$, period 2=$\{d_y+1,\dots,2d_y\}$ last period =$\lfloor T/(d_y+1) \rfloor (d_y+1),\dots,T$. In each period, the gradient is estimated by finite-difference exactly once at the first epoch.
Let $n$ denote the number of period, which is also the gradient estimation times. Then let $b_n:=\mathbb{E} \left( \|\tilde{y}_n-y^*(B)\|_2^2\right)$. Notice that $\Pi_{\mathcal{Y}} (y^*(B))=y^*(B)$ and $\|\Pi_{\mathcal{Y}} (\tilde{y}_{n+1})-\Pi_{\mathcal{Y}} (\tilde{y}_n)\|_2 \leq \|\tilde{y}_{n+1}-\tilde{y}_n\|_2$, then
\begin{equation}\label{eq:1}
\begin{split}
b_{n+1}
&= \mathbb{E} \left\{ \| \Pi_{\mathcal{Y}} (\tilde{y}_n+a_n G(\tilde{y}_n))-y^*(B)\|_2^2\right\}\\
&= \mathbb{E} \left\{ \| \Pi_{\mathcal{Y}} (\tilde{y}_n+a_n G(\tilde{y}_n))-\Pi_{\mathcal{Y}}(y^*(B))\|_2^2\right\}\\
&\leq \mathbb{E} \left\{\|\tilde{y}_n+a_n G(\tilde{y}_n) -y^*(B)\|_2^2 \right\}\\
&=b_n+a_n^2 \mathbb{E} \left( \|G(\tilde{y}_n)\|_2^2 \right) + 2 a_n \mathbb{E} \left[G(\tilde{y}_n)^T (\tilde{y}_n-y^*(B))\right]\\ 
\end{split}
\end{equation}
Let $g(\tilde{y}_n)=\frac{1}{c_n} \left([f(X_{n+1},y_{n}+c_n  e_1)-f(X_n,\tilde{y}_n)],\cdots,[f(X_{n+d},y_{n}+c_n e_{d_y})-f(X_n,\tilde{y}_n)]\right)^T$ and $g_B(\tilde{y}_n)=\frac{1}{c_n} \left([f_B(y_{n}+c_n e_1)-f_B(\tilde{y}_n)],\cdots,[f_B(y_{n}+c_n e_{d_y})-f_B(\tilde{y}_n)]\right)^T$.
Then, 
\begin{equation}\label{eq:2}
\begin{split}
\mathbb{E}\left[ G(\tilde{y}_n)^T (\tilde{y}_n-y^*(B))\right]
&=\mathbb{E} \left\{ \mathbb{E} \left[G(\tilde{y}_n)^T (\tilde{y}_n-y^*(B)) | \tilde{y}_n,X_n,X_{n+1},\ldots,X_{n+d_y} \in B\right]\right\}\\
&=\mathbb{E} \left\{ \mathbb{E} \left[ g(\tilde{y}_n)^T |X_n,X_{n+1},\ldots,X_{n+d_y} \in B\right] (\tilde{y}_n-y^*(B))\right\}\\
&=\mathbb{E} \left[ g_B(\tilde{y}_n)^T  (\tilde{y}_n-y^*(B))\right]\\
&=\mathbb{E} \left[\nabla f_B(\tilde{y}_n)^T (\tilde{y}_n-y^*(B))\right] + \mathbb{E} \left[ (g_B(\tilde{y}_n)-\nabla f_B(\tilde{y}_n))^T (\tilde{y}_n-y^*(B))\right]\\
\end{split}
\end{equation}
The first equality follows from tower law, second from the definition of $F(x,y,\xi)$ and third from the definition of $f_B(y)$.
According to strong concavity property of $f_B(y)$ (Lemma~\ref{Lemma:3}(2)), we have 
\begin{equation*}
\begin{split}
&f_B(\tilde{y}_n) \leq f_B(y^*(B))+ (\tilde{y}_n-y^*(B))^T \nabla f_B(y^*(B)) -\frac{1}{2} M_1 \|\tilde{y}_n-y^*(B)\|^2_2\\
&f_B(y^*(B)) \leq f_B(\tilde{y}_n) + (y^*(B)-\tilde{y}_n)^T \nabla f_B(\tilde{y}_n) -\frac{1}{2} M_1 
\|\tilde{y}_n-y^*(B)\|^2_2\\
\end{split}
\end{equation*}
Add them together,
\begin{equation*}
(\tilde{y}_n-y^*(B))^T (\nabla f_B(\tilde{y}_n)-\nabla f_B(y^*(B))) \leq -M_1 \|\tilde{y}_n-y^*(B)\|^2_2
\end{equation*}
Note that strong concavity of $f_B(y)$ implies that maximizer $y^*(B)$ is unique. By optimality of $y^*(B)$, we have 
\begin{equation*}
(\tilde{y}_n-y^*(B))^T \nabla f_B(y^*(B)) \leq 0
\end{equation*}
which together with last equation implies that $(\tilde{y}_n-y^*(B))^T \nabla f_B(\tilde{y}_n) \leq -M_1 \|\tilde{y}_n-y^*(B)\|^2_2$. Taking expectation of both sides,
\begin{equation}\label{eq:3}
\mathbb{E} \left[ \nabla f_B(\tilde{y}_n)^T (\tilde{y}_n-y^*(B)) \right] \leq -M_1 b_n
\end{equation}
Then by Lemma~\ref{Lemma:3}(5),
\begin{equation*}
\frac{1}{2} M_5 c_n \pmb{1} \geq g_B(\tilde{y}_n)-\nabla f_B(\tilde{y}_n) =\frac{1}{2} c_n \left(e_1^T \frac{d^2 }{d y^2} f_B(\eta_1) e_1, \cdots, e_d^T \frac{d^2 }{d y^2} f_B(\eta_d) e_d\right) \geq -\frac{1}{2} M_5 c_n \pmb{1}
\end{equation*}
Then by Cauthy-Schwarz inequality,
\begin{equation*}
\mathbb{E} \left[ (g_B(\tilde{y}_n)-\nabla f_B(\tilde{y}_n))^T (\tilde{y}_n-y^*(B))\right] \leq \frac{1}{2} M_5 c_n \mathbb{E} \left( \|\tilde{y}_n-y^*(B)\|_2\right) \leq \frac{1}{2} M_5 c_n \sqrt{b_n}
\end{equation*}
Thus,
\begin{equation}\label{eq:4}
\mathbb{E} \left[G(\tilde{y}_n)^T (\tilde{y}_n-y^*(B))\right] \leq -M_1 b_n +\frac{1}{2} M_5 c_n \sqrt{b_n}
\end{equation}
Furthermore, by Assumption 5,
\begin{equation}\label{eq:5}
\mathbb{E} \left( \| G(\tilde{y}_n)\|_2^2\right) \leq \frac{4 d_y M_3^2}{c_n^2}
\end{equation}
Therefore, by equations (~\ref{eq:1}),(~\ref{eq:2}),(~\ref{eq:3}),(~\ref{eq:4}),(~\ref{eq:5}),
\begin{equation*}
b_{n+1} \leq (1-2a_n M_1) b_n + a_n c_n M_5 \sqrt{b_n} + \frac{4 d_y M_3^2 a_n^2}{c_n^2}
\end{equation*}
Suppose that $a_n=a n^{-1}$ and $c_n=\delta n^{-\frac{1}{4}}$ with $1/(4 M_1) < a < 1/(2 M_1)$ and $\delta >0 $, we have 
\begin{equation*}
b_{n+1} \leq (1-\frac{2 a c}{n}) b_n + a \delta M_5 n^{-\frac{5}{4}} \sqrt{b_n} + \frac{4 d_y a^2 M_3^2 }{\delta^2} n^{-\frac{3}{2}}
\end{equation*}
Let $\alpha=2a M_1,\beta=\alpha \delta M_5,\omega=\frac{4d_y a^2 M_3^2}{\delta^2}$, by induction (see Lemma~\ref{Lemma:4}), there exists $\lambda >0 $ such that 
\begin{equation*}
b_n \leq \lambda n^{-\frac{1}{2}}
\end{equation*}
In each period, $d_y+1$ arms, $\tilde{y}_n, \tilde{y}_{n+1}, \ldots, y_{n+d_y}$ required to be implemented.
Notice that, by Lemma~\ref{Lemma:3}(6), $\nabla f_B (y^*(B))=0$ and Lemma~\ref{Lemma:3}(5), $\|\frac{d^2 }{d y^2} f_B(y)\|_2 \leq M_5$. Then by Taylor expansion,
\begin{equation*}
f_B(y^*(B))-f_B(\tilde{y}_n) \leq \frac{1}{2} M_5 \|\tilde{y}_n-y^*(B)\|_2^2
\end{equation*}
Taking expectation of both sides,
\begin{equation*}
\mathbb{E} [f_B(y^*(B))-f_B(\tilde{y}_n)] \leq \frac{1}{2} M_5 \mathbb{E} \left[\|\tilde{y}_n-y^*(B)\|_2^2\right]=\frac{1}{2} M_5 \lambda n^{-\frac{1}{2}}
\end{equation*}
and for $i=1,\ldots,d$,
\begin{equation*}
\begin{split}
\mathbb{E} [f_B(y^*(B))-f_B(\tilde{y}_n+ c_n e_i)] 
&\leq \frac{1}{2} M_5 \mathbb{E}  \left[ \|\tilde{y}_n+c_n e_i -y^*(B)\|_2^2\right] \\
&\leq M_5 \mathbb{E} \left(\|\tilde{y}_n-y^*(B)\|_2^2+c_n^2\right) \\
&\leq M_5 (\lambda + \delta^2) n^{-\frac{1}{2}}  
\end{split}
\end{equation*}

Recall that $\lambda$ is an affine function of $d_y$, then there exist a function $Q(d_y)$ such that
\begin{equation*}
\mathbb{E} [f_B(y^*(B))-f_B(y_t)] \leq \frac{\sqrt{d_y} Q(d_y)}{\sqrt{t}}
\end{equation*}
for every epoch $t$. $\quad \square \\$

\subsection{Proof of Theorem 1.}
\begin{theorem}
	For any function $f$ satisfying the assumptions in Section 2.1, the regret by Algorithm 1 is bounded by
	\begin{equation}
	R(T) \leq C d_y^{\frac{\alpha(d_x+6)}{2(d_x+2\alpha)}}T^{\frac{d_x+\alpha}{d_x+2\alpha}}
	\end{equation}
	for a constant $C$ that is independent of $d_x,d_y,T$.
\end{theorem}
We first prove a continuity result of maximizers in Lemma~~\ref{Lemma:5}. It gives an error bound for the distance between the maximizer in one bin and optimum for a covariate. The error will disappear as the diameter of $B$ shrinks to zero.

\begin{lemma}[H\"{o}lder continuous of $y^*(x)$ and $y^*(B)$]\label{Lemma:5} For a hypercube $B \subset \mathcal{Y}$, the diameter of arms space $\mathcal{Y}$ is $\sqrt{d_y}$ and let $d_B$ be the diameter of bin $B$. By Assumptions 1-4, there exists a uniform constant $M_6>0$ such that $\|y^*(B)-y^*(x)\|_2 \leq M_6 d_B^{\alpha/2}=M_6 d_{y}^{\alpha/4} d_x^{\alpha/4} K^{-\alpha/2}$ for any $x \in B$.
\end{lemma}

\textbf{Proof of Lemma 5.}
From twice continuously differentiable and strongly concavity of function $f$, we have that 
\begin{equation*}
\begin{split}
& \frac{M_1}{2} \|y^*(x)-y^*(B)\|_2^2 \leq f(x,y^*(x))-f(x,y^*(B)) \leq \frac{M_5}{2} \|y^*(x)-y^*(B)\|_2^2 \\
& \frac{M_1}{2} \|y^*(x)-y^*(B)\|_2^2 \leq f_B(y^*(B))-f_B(y^*(x)) \leq \frac{M_5}{2} \|y^*(x)-y^*(B)\|_2^2 \\
\end{split}
\end{equation*} 
Then add them together,
\begin{equation*}
\begin{split}
M_1 \|y^*(x)-y^*(B)\|_2^2 
& \leq f(x,y^*(x))-f_B(y^*(x))+f_B(y^*(B))-f(x,y^*(B)) \\
& = \mathbb{E} \left[ f(x,y^*(x))-f(X,y^*(x)) \Big| X \in B\right]\\
& \quad +\mathbb{E} \left[f(X,y^*(B)) -f(x,y^*(B))\Big| X \in B\right]\\
& \leq 2 \mathbb{E} \left[ M_2 \|x-X\|_2^{\alpha} | X \in B\right]\\
& \leq 2 M_2 d_B^{\alpha}\\
\end{split}
\end{equation*}
where $d_B= \sqrt{d_y d_x} /K$. So we get the conclusion,
\begin{equation*}
\|y^*(x)-y^*(B)\|_2 \leq M_6 d_B^{\alpha/2}=M_6 d_{y}^{\alpha/4} d_x^{\alpha/4} K^{-\alpha/2}
\end{equation*}

\textbf{Proof of Theorem 1.}
According to the algorithm, $\mathcal{B}_K$ denotes the partition formed by the bins. The regret associated with $X_t$ can be counted by bins $B \in \mathcal{B}_K$ into which $X_t$ falls. Therefore,
\begin{equation*}
\begin{split}
R(T)
&=\mathbb{E} \left[\sum_{t=1}^T \left( f^*(X_t)-f(X_t,\pi_t)\right)\right]=\mathbb{E} \left[\sum_{t=1}^T \sum_{B \in \mathcal{B}_K}\left( f^*(X_t)-f(X_t,\pi_t)\right) \mathbb{I}_{\{X_t \in B\}}\right]\\
\end{split}
\end{equation*}

According to Lemma~\ref{Lemma:1}(3), we have $f^*(X_t)-f(X_t,y_t) \leq \frac{1}{2} M_5 \|y_t-y^*(X_t)\|_2^2$. The distance between $y_t$ and $y^*(X_t)$ can be bounded by $\|y_t-y^*(B)\|_2$ and $\|y^*(B)-y^*(X_t)\|_2$, where the first term is bounded by the error bound of stochastic approximation proved in Proposition 1 and the second term is bounded by the diameter of bin $B$ using Lemma~\ref{Lemma:5}. Therefore,
\begin{equation*}
\begin{split}
\mathbb{E} \left[\left(f^*(X_t)-f^*(X_t,y_t)\right) \mathbb{I}_{\{X_t \in B \}} \right] 
&\leq \frac{1}{2} M_5 \mathbb{E} \left[\|y_t-y^*(X_t)\|^2_2 \mathbb{I}_{\{X_t \in B \}}\right]\\
&\leq M_5 \left\{ \mathbb{E} \left[\|y_t-y^*(B)\|^2_2 \right]+\mathbb{E} \left[\|y^*(B)-y^*(X_t)\|^2_2 \right] \right\}\\
&\leq M_5 \mathbb{E} \left[ \|y_t-y^*(B)\|_2^2 \right]+M_5  \left( M_6 d_{y}^{\alpha/4} d_x^{\alpha/4} K^{-\alpha/2}\right)^2\\
&\leq M_5 \left[\frac{Q(d_y)}{\sqrt{t_B/(d_y+1)}}\right] + M_5 M_6^2 d_{y}^{\alpha/2} d_x^{\alpha/2} K^{-\alpha}\\
\end{split}
\end{equation*}
where $Q(d_y)$ is the function defined in Proposition 1 and $t_B$ denotes observation times in $B$. After deriving the regret bound in one period, we can sum them together and obtain the bound of total regret. 

Since $B_K$ forms a partition of the covariate space and $X$ always falls into one of the bins. The worst case is that all the covariates are uniformly distributed in the whole covariate space, for the regret incurred in each bin increases in the reciprocal order as observations increasing. Thus the total regret is less than the accumulative regret in $K^{d_x}$ bins when covariates occur $T/K^{d_x}$ times in each bin.
\begin{equation*}
\begin{split}
R(T) 
& \leq K^{d_x} (d_y+1)M_5 \sum_{t=1}^{\left\lceil T/ \left[(d_y+1) K^{d_x}\right] \right\rceil}  \left[Q_{d_y}/\sqrt{t}\right]+T M_5 M_6^2 d_{y}^{\alpha/2} d_x^{\alpha/2} K^{-\alpha}\\
\end{split}
\end{equation*}
By the summation of series, $\sum_{t=1}^{n} 1/\sqrt{t} \leq 1+2 \sqrt{n}$, we have 
\begin{equation*}
\begin{split}
\frac{R(T)}{M_5} 
& \leq K^{d_x} (d_y+1) Q_{d_y} \sqrt{1+\frac{T}{(d_y+1) K^{d_x}}} +T M_5 M_6^2 d_{y}^{\alpha/2} d_x^{\alpha/2} K^{-\alpha}\\
\end{split}
\end{equation*}
Using the Cauchy inequality,
\begin{equation*}
\begin{split}
\frac{R(T)}{M_5} 
& \leq K^{d_x} (d_y+1) Q_{d_y} + \sqrt{d_y+1} K^{d_x/2} Q_{d_y} \sqrt{T} +T M_5 M_6^2 d_{y}^{\alpha/2} d_x^{\alpha/2} K^{-\alpha}
\end{split}
\end{equation*}

We separate in 2 cases to design the $K$ that minimizes the total regret.
\begin{description}
	\item[(1)] If $T \geq K^{d_x}$, then  
	\begin{equation*}
	\frac{R(T)}{M_5}  \leq  b_0 d_y^{3/2} K^{d_x/2} \sqrt{T} + b_1 d_{y}^{\alpha/2} d_x^{\alpha/2} K^{-\alpha} T
	\end{equation*}
	Hence, to minimize the total regret, by the choice of $K=\left(d_x^{\alpha-2} d_y^{\alpha-3} T\right)^{\frac{1}{d_x+2\alpha}}$ (satisfying $T \geq K^{d_x}$), 
	\begin{equation*}
	R(T) \leq c_2 d_x^{\frac{(\alpha-2) d_x}{2 (d_x +2 \alpha)}} d_y^{\frac{\alpha(d_x+6)}{2(d_x+2\alpha)}}T^{\frac{d_x+\alpha}{d_x+2\alpha}}
	\end{equation*}
	\item[(2)] If $T \leq K^{d_x}$, then
	\begin{equation*}
	\frac{R(T)}{M_5} \leq b_3 d_y^2  T + b_4 d_x^{\alpha/2} K^{-\alpha} T
	\end{equation*}
	Hence,
	\begin{equation*}
	R(T) \leq b_5 T
	\end{equation*}
	
\end{description}

Hence, we choose $K=\left(d_x^{\alpha-2} d_y^{\alpha-3} T\right)^{\frac{1}{d_x+2\alpha}}$ and complete the proof of Theorem 1. $\Box$

\subsection{Proof of Theorem 2.}
The adaptive algorithm is more challenging to analyze. That's mainly because it split bins adaptively which cause expected randomness in determining the size of the bins. In the algorithm, we keep a dynamic partition $\mathcal{B}_t$ of the covariate space consisting of all the bins in each period $t$. The partition is mutually exclusive and collectively exhaustive. Specifically, 
\begin{equation*}
\mathcal{B}_t=\bigcup_{l=0}^L \left\{B:l(B)=l\right\}
\end{equation*}
where $l(B)$ denotes the level of bin $B$. The partition $\mathcal{B}_t$ will be gradually refined in the algorithm and updated at the end of each period.

Similar to the proof of Theorem 1, we first extend Lemma~\ref{Lemma:5} to a result in bin $B$ of level $l$.

\begin{lemma}[Extension of Lemma~\ref{Lemma:5}]\label{Lemma:6} For a bin $B$ in level $l$, the diameter of arms space $\mathcal{Y}$ is $\sqrt{d_y}$ and let $d_B$ be the diameter of bin $B$. By Assumptions 1-4, there exists a uniform constant $M_6>0$ such that $\|y^*(B)-y^*(x)\|_2 \leq M_6 d_B^{\alpha/2}=M_6 d_{y}^{\alpha/4} d_x^{\alpha/4} 2^{-\alpha l/2}$ for any $x \in B$.
\end{lemma}

\textbf{Proof of Lemma 6}. 
In the bin $B$ of level $l$, the diameter $d_B=\sqrt{d_y d_x} 2^{-l}$. So following the proof of Lemma~\ref{Lemma:5}, we have 
\begin{equation*}
\|y^*(x)-y^*(B)\|_2 \leq M_6 d_B^{\alpha/2} = M_6 d_{y}^{\alpha/4} d_x^{\alpha/4} 2^{-\alpha l/2}
\end{equation*}

Proposition 1 provides a finite-time convergence result of algorithm 1. It only gives an error bound in a single bin. So we generalize it to be applicable to the bins in different levels.

\begin{proposition}
	In a bin $B$ of level $l$, let $a_n=a n^{-1}$ and $c_n=\delta n^{-\frac{1}{4}}$, where $1/(4M_1)<a<1/(2M_1)$ and $\delta>0$.
	Then the regret of Algorithm 1 in bin $B$ satisfies
	\begin{equation*}\label{eq:proposition2}
	\begin{split}
	& \mathbb{E} \left[f_B(y^\ast(B))-f_B(y_t)\right] 
	\leq \frac{\mathbb{E} [Q_B(d_y)]}{\sqrt{t/(d_y+1)}} \\
	& \leq \frac{1}{\sqrt{t/(d_y+1)}} \left\{\left(\frac{2 M_5}{\sqrt{n}}\right)^{l-1} q_1 +\frac{ 2^{\alpha} \sqrt{n}}{\sqrt{n}-2^{1+\alpha} M_5} c_1  2^{- \alpha l}+\frac{\sqrt{n}}{\sqrt{n}-2 M_5} c_0\right\} \\
	\end{split} 
	\end{equation*}
	where $q_1=\|\tilde{y}_0-y^*(B_{0})\|_2^2$ is the initial error in  bin $B_{0}$.
\end{proposition}

\textbf{Proof of Proposition 2}.

In the result of Proposition 1, we know that
\begin{equation*}
\begin{split}
Q_B(d_y)=
& M_5 \delta^2 + M_5 \max \left\{ \|\tilde{y}_0-y^*(B)\|_2^2, \right.\\
& \frac{1}{(4 \delta M_5 -1 )^2} \left. \left(2 \delta M_5 + \sqrt{4 \delta^2 M_5^2+8 d_y a^2 M_7^2 (4a M_1 -1)/\delta^2}\right)^2 \right\}\\
\leq 
& M_5 \delta^2 + M_5 \|\tilde{y}_0-y^*(B)\|_2^2 \\
& \ +\frac{M_5}{(4 \delta M_5 -1 )^2}  \left(2 \delta M_5 + \sqrt{4 \delta^2 M_5^2+8 d_y a^2 M_7^2 (4a M_1 -1)/\delta^2}\right)^2 \\
\triangleq 
& M_5 \|\tilde{y}_0-y^*(B)\|_2^2+ c_0 (d_y) \\
\end{split}
\end{equation*}
where $c_0 (d_y)$ is a linear function of $d_y$.

At the right-hand side, $\tilde{y}_0$ is the initial solution when context is observed in bin $B$. According to the Algorithm 3, it's a random variable defined as the last solution of the parent bin of $B$. So $Q_B(d_y)$ depends on the information before the bin $B$ is generated. Let $Pa(B)$ denotes the parent bin of $B$, so the level of $Pa(B)$ is $l-1$. Then, let $y^*_{Pa(B)}$ denotes the last solution in $Pa(B)$ and $y^*(Pa(B))$ denotes the true optimum of $Pa(B)$.

\begin{equation*}
\begin{split}
\mathbb{E} \left[Q_B(d_y)\right] 
& \leq  M_5 \mathbb{E} \left[\|\tilde{y}_0-y^*(B)\|_2^2 \right] + c_0 (d_y) \\
& = M_5 \mathbb{E} \left[\|y^*_{Pa(B)}-y^*(B)\|_2^2 \right] + c_0 (d_y) \\
& \leq 2 M_5 \mathbb{E} \left[\|y^*_{Pa(B)}-y^*(Pa(B))\|_2^2 \right]+2 M_5 \mathbb{E} \left[\|y^*(Pa(B))-y^*(B)\|_2^2 \right]+ c_0 (d_y) \\ 
& \leq \frac{2 M_5}{\sqrt{n_{l-1} /(d_y+1)}} \mathbb{E} \left[Q_{Pa(B)}(d_y)\right] + 2 M_5 M_6^2 d_{y}^{\alpha/2} d_x^{\alpha/2} 2^{-\alpha (l-1)} + c_0 (d_y)
\end{split}
\end{equation*}

The last inequality follows from applying Proposition 1 to bin $Pa(B)$ and Lemma~\ref{Lemma:6}. 

Since all the bins in the same level are constructed in the same procedure, $\mathbb{E} \left[Q_B(d_y)\right]$ are all equal. Let $q_l= \mathbb{E} \left[Q_B(d_y)\right]$, then $q_{l-1}=\mathbb{E} \left[Q_{Pa(B)}(d_y)\right]$. Thus, we have shown that
\begin{equation*}
q_l \leq \frac{2 M_5}{\sqrt{n_{l-1} /(d_y+1)}} q_{l-1} + c_1 2^{-\alpha (l-1)} + c_0
\end{equation*}
where $c_1 \triangleq 2 M_5 M_6^2 d_{y}^{\alpha/2} d_x^{\alpha/2} $ and $c_0=c_0(d_y)$. According to the main part, $n_l=n(d_y+1)$. 
Therefore, the initial error $q_l$ forms a sequence.
\begin{equation*}
\begin{split}
q_l 
& \leq \frac{2 M_5}{\sqrt{n}} q_{l-1} + c_1  2^{- \alpha (l-1)}+c_0 \\
&\leq \left(\frac{2 M_5}{\sqrt{n}}\right)^{l-1} q_1 + \sum_{i=0}^{l-2} \left(\frac{2 M_5}{\sqrt{n}}\right)^i 2^{- \alpha (l-1-i)} c_1  + \sum_{i=0}^{l-2} \left(\frac{2 M_5}{\sqrt{n}}\right)^i c_0\\
&= \left(\frac{2 M_5}{\sqrt{n}}\right)^{l-1} q_1 + 2^{- \alpha (l-1)} \dfrac{1-(2^{1+\alpha} M_5/\sqrt{n})^{l-1}}{1-2^{1+\alpha} M_5/\sqrt{n}} c_1  + \frac{1-(2 M_5/\sqrt{n})^{l-1}}{1-2 M_5/\sqrt{n}} c_0\\
& \leq \left(\frac{2 M_5}{\sqrt{n}}\right)^{l-1} q_1 + 2^{- \alpha (l-1)} \dfrac{1}{1-2^{1+\alpha} M_5/\sqrt{n}} c_1  + \frac{1}{1-2 M_5/\sqrt{n}} c_0\\
& =\left(\frac{2 M_5}{\sqrt{n}}\right)^{l-1} q_1 +\frac{ 2^{\alpha} \sqrt{n}}{\sqrt{n}-2^{1+\alpha} M_5} c_1  2^{- \alpha l}+\frac{\sqrt{n}}{\sqrt{n}-2 M_5} c_0 
\end{split} 
\end{equation*} 

where $q_1=\|\tilde{y}_0-y^*(B_{0})\|_2^2$ is the initial error in  bin $B_{0}$.

Thus, we conclude to prove Proposition~\ref{eq:proposition2}.
\begin{equation*}
\begin{split}
& \mathbb{E} \left[f_B(y^\ast(B))-f_B(y_t)\right] 
\leq \frac{\mathbb{E} [Q_B(d_y)]}{\sqrt{t/(d_y+1)}} \\
& \leq \frac{1}{\sqrt{t/(d_y+1)}} \left\{\left(\frac{2 M_5}{\sqrt{n}}\right)^{l-1} q_1 +\frac{ 2^{\alpha} \sqrt{n}}{\sqrt{n}-2^{1+\alpha} M_5} c_1  2^{- \alpha l}+\frac{\sqrt{n}}{\sqrt{n}-2 M_5} c_0\right\} \ \ \square\\
\end{split} 
\end{equation*}

\textbf{Proof of Theorem 2.} According to the algorithm, $\mathcal{B}_t$ denotes the partition formed by the bins at time t when $X_t$ is generated. The regret associated with $X_t$ can be counted by bins $B \in \mathcal{B}_t$ into which $X_t$ falls. Meanwhile, the level of $B$ is at most $K$. Therefore,
\begin{equation*}
\begin{split}
R(T)
&=\mathbb{E} \left[\sum_{t=1}^T \left( f^*(X_t)-f(X_t,y_t)\right)\right]=\mathbb{E} \left[\sum_{t=1}^T \sum_{B \in \mathcal{B}_t}\left( f^*(X_t)-f(X_t,y_t)\right) \mathbb{I}_{\{X_t \in B\}}\right]\\
&=\mathbb{E} \left[\sum_{t=1}^T \sum_{l=0}^L \sum_{\{B:l(B)=l \}}\left( f^*(X_t)-f(X_t,y_t)\right) \mathbb{I}_{\{X_t \in B ,B \in \mathcal{B}_t\}}\right]\\
\end{split}
\end{equation*}

According to Lemma~\ref{Lemma:1}(3), we have $f^*(X_t)-f(X_t,y_t) \leq \frac{1}{2} M_5 \|y_t-y^*(X_t)\|_2^2$. The distance between $y_t$ and $y^*(X_t)$ can be bounded by $\|y_t-y^*(B)\|_2$ and $\|y^*(B)-y^*(X_t)\|_2$, where the first term is bounded by the error bound of stochastic approximation proved in Proposition~\ref{eq:proposition2} and the second term is bounded by the diameter of bin $B$ using Lemma~\ref{Lemma:5}. Therefore,

\begin{equation*}
\begin{split}
\mathbb{E} \left[\left(f^*(X_t)-f^*(X_t,y_t)\right) \mathbb{I}_{\{X_t \in B \}} \right] 
&\leq \frac{1}{2} M_5 \mathbb{E} \left[\|y_t-y^*(X_t)\|^2_2 \mathbb{I}_{\{X_t \in B \}}\right]\\
&\leq M_5 \left\{ \mathbb{E} \left[\|y_t-y^*(B)\|^2_2 \right]+\mathbb{E} \left[\|y^*(B)-y^*(X_t)\|^2_2 \right] \right\}\\
&\leq M_5 \mathbb{E} \left[ \|y_t-y^*(B)\|_2^2 \right]+M_5  \left( M_6 d_{y}^{\alpha/4} d_x^{\alpha/4} K^{-\alpha/2}\right)^2\\
&\leq M_5 \left[\frac{\mathbb{E} [Q_B(d_y)]}{\sqrt{t_B/(d_y+1)}}\right] + M_5 M_6^2 d_{y}^{\alpha/2} d_x^{\alpha/2} K^{-\alpha}\\
&\leq M_5 \frac{1}{\sqrt{t_B/(d_y+1)}} \left\{\left(\frac{2 M_5}{\sqrt{n}}\right)^{l-1} q_1 +\frac{ 2^{\alpha} \sqrt{n}}{\sqrt{n}-2^{1+\alpha} M_5} c_1  2^{- \alpha l}+\frac{\sqrt{n}}{\sqrt{n}-2 M_5} c_0 \right\}\\
&+ M_5 M_6^2 d_{y}^{\alpha/2} d_x^{\alpha/2} K^{-\alpha}\\
\end{split}
\end{equation*}

After deriving the regret bound in one period, we can sum them together and obtain the bound of total regret. 
\begin{equation*}
\begin{split}
R(T)
& = \mathbb{E} \left[\sum_{t=1}^T \sum_{l=0}^L \sum_{\{B:l(B)=l\}}\left( f^*(X_t)-f(X_t,y_t)\right) \mathbb{I}_{\{X_t \in B ,B \in \mathcal{B}_t\}}\right]\\
& \leq \mathbb{E} \left[\sum_{l=0}^{L-1} \sum_{\{B:l(B)=l\}} \sum_{t=1}^T \left( f^*(X_t)-f(X_t,y_t)\right) \mathbb{I}_{\{X_t \in B ,B \in \mathcal{B}_t\}}\right]\\
&+\sum_{\{B:l(B)=L\}} \sum_{t=1}^T \mathbb{E} \left[\left( f^*(X_t)-f(X_t,y_t)\right) \right]  \\
\end{split}
\end{equation*}

For the first term, note that $\{X_t \in B ,B \in \mathcal{B}_t\}$ occurs for at most $n_l$ times for given $B$ with $l(B)=l$. Moreover, there are $2^{dl}$ bins with level $l$ except for $L$. Total regret incurred in the first $L-1$ levels is bounded by the regret incurred in full tree of level $L-1$. 

For the second term, $\{B:l(B)=L\}$ forms a partition of the covariate space and $X$ always falls into one of the bins. The worst case is that all the covariates are uniformly distributed in the whole covariate space, for the regret incurred in each bin increases in the reciprocal order as observations increasing. Thus the total regret incurred in the last level is less than the accumulative regret in $2^{d_xL}$ bins when covariates occur $T/2^{d_xL}$ times in each bin.
\begin{equation*}
\begin{split}
R(T) 
& \leq \sum_{l=0}^{L-1} 2^{d_xl} \left\{ \sum_{i=1}^{n} (d_y+1)L \mathbb{E} \left[Q_{Bl}  /\sqrt{i} \right] + n M_5 M_7^2 d_x 2^{-\alpha l}\right\} \\
& \quad + 2^{d_x L} (d_y+1)L \sum_{t=1}^{\left\lceil T/ \left[(d_y+1) 2^{d_xL}\right] \right\rceil} \mathbb{E} \left[Q_{B_L}/\sqrt{t}\right]+T M_5 M_7^2 d_x 2^{- \alpha L}\\
\end{split}
\end{equation*}
By the summation of series, $\sum_{i=1}^{n} 1/\sqrt{i} \leq 1+2 \sqrt{n}$, we have 
\begin{equation*}
\begin{split}
\frac{R(T)}{M_5} 
& \leq \sum_{l=0}^{L-1} 2^{d_xl} \left\{ (d_y+1)(1+2\sqrt{n}) \mathbb{E} \left[Q_{B_l}  \right] + n M_7^2 d_x 2^{- \alpha l}\right\} \\
& + 2^{d_x L+1}(d_y+1)\sqrt{1+\frac{T}{(d_y+1) 2^{d_xL}}} \mathbb{E} \left[Q_{B_L}\right]+T M_7^2 d_x 2^{-\alpha L}\\
\end{split}
\end{equation*}
Replace $\mathbb{E} \left[Q_{B_l}\right]$ using Proposition~\ref{eq:proposition2},
\begin{equation*}
\begin{split}
\frac{R(T)}{M_5}
& \leq \sum_{l=0}^{L-1} 2^{d_xl} \left\{ (d_y+1)(1+2\sqrt{n}) \left[\left(\frac{2 M_5}{\sqrt{n}}\right)^{l-1} q_1 +\frac{ 2^{\alpha} \sqrt{n}}{\sqrt{n}-2^{1+\alpha} M_5} c_1  2^{- \alpha l}+\frac{\sqrt{n}}{\sqrt{n}-2 M_5} c_0\right]+ n M_7^2 d_x 2^{- \alpha l}\right\} \\
&\quad + 2 \sqrt{(d_y+1) 2^{d_xL} T+(d_y+1)^2 2^{2d_xL}} \left[\left(\frac{2 M_5}{\sqrt{n}}\right)^{l-1} q_1 +\frac{ 2^{\alpha} \sqrt{n}}{\sqrt{n}-2^{1+\alpha} M_5} c_1  2^{- \alpha l}+\frac{\sqrt{n}}{\sqrt{n}-2 M_5} c_0\right]\\
& \quad +T M_7^2 d_x 2^{-\alpha L}\\
\end{split}
\end{equation*}
Recall that $n> \max\{64 M_5^2,2^{1+\alpha}\}$, then $2 M_5 /\sqrt{n} <1 $ and $2^{d_x+1} M_5/\sqrt{n} < 2^{d_x}$. Hence, can be further simplified as
\begin{equation*}
\begin{split}
\frac{R(T)}{M_5}
& = \frac{\sqrt{n}}{2 M_5} (d_y+1)(1+2\sqrt{n}) q_1 \left[\sum_{l=0}^{L-1} (2^{d_x+1} M_5/\sqrt{n})^l\right] +\frac{\sqrt{n}}{\sqrt{n}-2 M_5} (d_y+1)(1+2 \sqrt{n}) c_0 \left[\sum_{l=0}^{L-1} 2^{d_xl} \right] \\
&\quad + \left[\frac{2^{\alpha}\sqrt{n}}{\sqrt{n}-2^{1+\alpha} M_5} (d_y+1)(1+2 \sqrt{n}) c_1 d_x+n M^2_3 d_x\right] \left[ \sum_{l=0}^{L-1} (2^{d_x-\alpha})^l \right]\\
&\quad +c_4 d_y^{\frac{3}{2}} \sqrt{T} \sqrt{2^{d_xL}} +c_4 d_y^2 2^{d_xL}+ T M_7^2 d 2^{-\alpha L}\\
& \leq c_2 d_y^2 \left[\sum_{l=0}^{L-1} 2^{d_xl}\right] +c_3 d_x d_y \left[\sum_{l=0}^{L-1} (2^{d_x-\alpha})^l\right] +c_4 d_y^{\frac{3}{2}} \sqrt{T} \sqrt{2^{d_xL}} +c_4 d_y^2 2^{d_xL} + c_5 d_xT 2^{-\alpha L}\\
& \leq c_2 d_y^2  2^{d_x L} + c_3 d_x d_y 2^{(d_x-\alpha)L} +c_4 d_y^{\frac{3}{2}} \sqrt{T} \sqrt{2^{d_xL}} +c_4 d_y^2 2^{d_xL} + c_5 T d_x 2^{-\alpha L}\\
\end{split}
\end{equation*}
Suppose $T$ is large enough that the level of tree $L>\frac{\log d_x+\log d_y}{2 \log 2}$, so the second term can be merged into the first term.
\begin{equation*}
\frac{R(T)}{M_5} \leq (c_2+c_3+c_4) d_y^2  2^{d_xL} +c_4 d_y^{\frac{3}{2}} \sqrt{T} \sqrt{2^{d_xL}} + c_5 T d_x 2^{-\alpha L}
\end{equation*}
We separate in 2 cases to obtain the $L$ that minimizes the total regret.
\begin{description}
	\item[(1)] If $d_xL \leq \log T$, then $T \geq 2^{d_xL}$, and  
	\begin{equation*}
	\frac{R(T)}{M_5}  \leq  c_6 d_y^{\frac{3}{2}} \sqrt{T} \sqrt{2^{d_xL}} + c_5 T d_x 2^{-\alpha L}
	\end{equation*}
	Hence, to minimize the total regret, by the choice of $L=\dfrac{\log T + 2 \log \alpha- 3 \log d_y + 2 \log d_x}{(d_x+2 \alpha) \log 2}$ (satisfying $d_xL \leq \log T$), 
	\begin{equation*}
	R(T) \leq c_7 \alpha^{\frac{d_x}{d_x+2\alpha}} d_y^{\frac{3}{2}+\frac{2d_x}{3(d_x+2\alpha)}} d_x^{\frac{d_x}{d_x+2\alpha}} T^{\frac{d_x+\alpha}{d_x+2\alpha}}
	\end{equation*}
	\item[(2)] If $d_xL > \log T$, then $T < 2^{d_xL}$, and 
	\begin{equation*}
	\frac{R(T)}{M_5} \leq (c_2+c_3+c_4) d_y^2  2^{d_xL} + c_5 d_x T 2^{-\alpha L}
	\end{equation*}
	Hence, to minimize the total regret, by the choice of $L=\dfrac{\log T+\log \alpha-2 \log d_y}{(d_x+\alpha) \log 2}$ (but $d_xL > \log T$), 
	\begin{equation*}
	R(T) \leq c_7 T
	\end{equation*}
	
\end{description}

Hence, we complete our proof of Theorem~\ref{eq:proposition2}. $\Box$

\subsection{Proof of Theroem 3.}

\begin{proposition}\label{prop:example_assum}
	We construct the example, for $x \in B_j$,
	\begin{equation*}
	\begin{split}
	f_{\omega}(x,y)
	&=-\|y\|_2^2 +2 \omega_j^T y d(x,\partial B_j)\\
	&=-y_{(1)}^2-y_{(2)}^2-\cdots -y_{(d_y)}^2 + 2 \omega_{(1,j)} y_{(1)} d(x,\partial B_j)+ 2 \omega_{(2,j)} y_{(2)} d(x,\partial B_j)+ \cdots + 2\omega_{(d_y,j)} y_{(d_y)} d(x,\partial B_j)\\
	\end{split}
	\end{equation*}
	The example satisfies assumptions.
\end{proposition}

\textbf{Proof of Proposition~\ref{prop:example_assum}}
\begin{description}
	\item[(1,2)] 
	The first order and second order derivative respect to $y$ are,
	\begin{equation*}
	\nabla_y f_{\omega} (x,y) = -2y+ 2 \omega_j d(x,\partial B_j), \  \nabla_y^2 f_{\omega} (x,y) = -2I
	\end{equation*}
	So the first and second assumptions are obviously satisfied.
	\item[(3)]
	\begin{equation*}
	\begin{split}
	f_B(y)
	&= \mathbb{E} [f(x,y)| x \in B]\\
	&= -\|y\|_2^2 + 2 \omega_j^T y \mathbb{E} \left[\sum_{j=1}^{M_d} \mathbb{I} \{X \in B_j\} d(X, \partial B_j) \Big| X \in B\right]\\
	&= -\|y\|_2^2 +2 y^T \sum_{j=1}^{M_d} \omega_j \mathbb{P} (B_j \cap B) \mathbb{E} \left[ d(X, \partial B_j) \Big| X \in B_j \cap B \right]\\
	\end{split}
	\end{equation*}
	Then,
	\begin{equation*}
	p^*(B)=\sum_{j=1}^{M_d} \omega_j \mathbb{P} (B_j \cap B) \mathbb{E} \left[ d(X, \partial B_j) \Big| X \in B_j \cap B \right]
	\end{equation*}
	is in the interior of the decision space.
	\item[(6)]
	Since the distance to boundary $d(x,\partial B_j)$ is a kind of infinity norm, its derivative to $x$ is less than 1. So the partial derivative 
	\begin{equation*}
	\Big| \frac{\partial f_{\omega} (x,y)}{\partial x_k \partial y_i} \Big| = \Big| 2 \omega_{(i,j)} \frac{\partial d(x,\partial B_j)}{\partial x_k} \Big| \leq 2
	\end{equation*}
	Assumption 6 is satisfied. 
	
\end{description}

\textbf{Proof of Theorem 3.}
\begin{equation*}
\begin{split}
\sup_{f \in  C} R_{\pi} 
&= \sup_{f \in C} \sum_{t=1}^T \mathbb{E} [f^*(X_t)-f(X_t,\pi_t)] \\
&=\sup_{f \in C} \sum_{t=1}^T \sum_{j=1}^{M^{d_x}} \mathbb{E} \left[(f^*(X_t)-f(X_t,\pi_t)) \mathbb{I} \{X_t \in B_j \}\right]\\
&\geq \frac{M_5}{2} \sup_{f \in  C} \sum_{t=1}^T \sum_{j=1}^{M^{d_x}} \mathbb{E} \left[\|y^*(X_t)-y_t\|_2^2 \mathbb{I} \{X_t \in B_j\}\right]\\
&\geq \frac{M_5}{2} 2^{-d_y M^{d_x}} \sum_{\omega} \sum_{t=1}^T \sum_{j=1}^{M^{d_x}} \mathbb{E}_{f_{\omega}}^{\pi} \left[\|y^*(X_t)-y_t\|_2^2 \mathbb{I} \{X_t \in B_j\}\right]\\
\end{split}
\end{equation*}
Where the total number of instances is ${C}=2^{d_y M^{d_x}}$. For a given bin $B_j$, we focus on $f_{\omega_{-j},\omega_j}$ which only differs for $x \in B_j$. We rewrite the summation $\sum_{\omega}$ as $\sum_{\omega_{-j} \in \{0,1\}^{d_y(M^{d_x}-1)}} \sum_{\omega_j \in \{0,1\}^{d_y}}$.
\begin{equation*}
\begin{split}
\sup_{f \in  C} R_{\pi} 
&\geq \frac{M_5}{2} 2^{-d_y M^{d_x}} \sum_{\omega_{-j} \in \{0,1\}^{d_y(M^{d_x}-1)}} \sum_{\omega_j \in \{0,1\}^{d_y}} \sum_{t=1}^T \sum_{j=1}^{M^{d_x}} \mathbb{E}_{f_{\omega}}^{\pi} \left[\|y^*(X_t)-y_t\|_2^2 \mathbb{I} \{X_t \in B_j\}\right]\\
&\geq \frac{M_5}{2} 2^{-d_y M^{d_x}} \sum_{\omega_{-j}} \sum_{i=1}^{d_y} \sum_{\omega_{(-i,j)}} \sum_{t=1}^T \sum_{j=1}^{M^{d_x}} \mathbb{E}_{f_{\omega_{-j},\omega_{\{-i,j\}}},0}^{\pi} \frac{num}{den}\left[\|y_t\|_2^2 \mathbb{I} \{X_t \in B_j\}\right]\\
\end{split}
\end{equation*}
Let $z_{\omega_{-j},\omega_{-i,j}}=1/(8M^2) \sum_{t=1}^T \mathbb{E}_{f_{\omega_{-j},\omega_{\{-i,j\}}},0}^{\pi} \left[\|y_t\|_2^2 \mathbb{I} \{X_t \in B_j\}\right]$, then
\begin{equation*}
\sup_{f \in  C} R_{\pi} \geq \frac{M_5}{2} \frac{8M^2}{2^{d_y M^{d_x}}} \sum_{j=1}^{M^{d_x}} \sum_{\omega_{-j}} \sum_{i=1}^{d_y} \sum_{\omega_{(-i,j)}} z_{\omega_{-j},\omega_{\{-i,j\}}}
\end{equation*}
We complete the other side by the information theory. $\mathbb{P} (X_t \in B_j)=M^{-d_x}$, then
\begin{equation*}
\begin{split}
\mathbb{E}^{\pi}_{f_\omega} \left[\|y^*(X_t)-y_t\|_2^2] \mathbb{I}_{X_t \in B_j}\right] 
&= \frac{1}{M^{d_x}} \mathbb{E}^{\pi}_{f_\omega} \left[\|y^*(X_t)-y_t\|_2^2 | X_t \in B_j \right]\\
&= \frac{1}{M^{d_x}} \mathbb{E}^{\pi}_{f_\omega} \left[\mathbb{E} \left[\|y^*(X_t)-y_t\|_2^2 | \mathcal{F}_{t-1},X_t \in B_j \right]\right]\\
\end{split} 
\end{equation*}

\begin{equation*}
\begin{split}
&\sum_{\omega_j \in \{0,1\}^{d_y}} \mathbb{E}^{\pi}_{f_{\omega_{-j},\omega_j}} \left[\|y^*(X_t)-y_t\|_2^2 \mathbb{I} \{X_t \in B_j\}\right]\\
&= \sum_{\omega_j} \mathbb{E}^{\pi}_{f_{\omega_{-j},\omega_j}} \left[\sum_{i=1}^{d_y} (y^*_{(i)}(X_t)-y_t^{(i)})^2 \mathbb{I} \{X_t \in B_j\}\right]\\
&=\frac{1}{M^{d_x}} \sum_{i=1}^{d_y} \sum_{\omega_j} \mathbb{E}^{\pi}_{f_{\omega_{-j},\omega_j}} \left\{ \mathbb{E} \left[\sum_{i=1}^{d_y} (y^*_{(i)}(X_t)-y_t^{(i)})^2 \Big| X_t \in B_j, \mathcal{F}_{t-1} \right] \right\}\\
& \geq \frac{1}{M^{d_x}} \sum_{i=1}^{d_y} \sum_{\omega_j} \frac{s^2}{M^2} \mathbb{E}^{\pi}_{f_{\omega_{-j},\omega_j}} \left[ \mathbb{P}_{X_t}^{B_j,t-1} \left(\|y_{(i)} (X_t)-y_t^{(i)}\| \geq \frac{s}{M}\right) \right]\\
& = \frac{s^2}{M^{d_x+2}} \sum_{i=1}^{d_y} \sum_{\omega_{\{-i,j\}}} \left( \mathbb{E}^{\pi}_{f_{\omega_{-j},\omega_{-i,j},0}} \left[ \mathbb{P}_{X_t}^{B_j,t-1} \left( |y_t^{(i)}| \geq \frac{s}{M}, A\right) \right] + \mathbb{E}^{\pi}_{f_{\omega_{-j},\omega_{-i,j},1}} \left[ \mathbb{P}_{X_t}^{B_j,t-1} \left(\|d(X_t,\partial B_j)-y_t^{(i)}\| \geq \frac{s}{M},A\right) \right]\right)\\
& \geq \frac{s^2}{M^{d_x+2}} (1-4s)^{d_x} \sum_{i=1}^{d_y} \sum_{\omega_{\{-i,j\}}} \left(\mathbb{P}^{\pi}_{f_{\omega_{-j},\omega_{\{-i,j\}},0}} \left(\Pi_t^{(i)}=1 | X_t \in A\right) + \mathbb{P}^{\pi}_{f_{\omega_{-j},\omega_{\{-i,j\}},1}} \left(\Pi_t^{(i)}=0 | X_t \in A\right)\right)\\
& \geq \frac{s^2}{M^{d_x+2}} \sum_{i=1}^{d_y} \frac{(1-4s)^{d_x}}{2} \sum_{\omega_{\{-i,j\}}} \exp\left(- \mathcal{K} \left(\mathbb{\mu}^{\pi,t-1}_{f_{\omega_{-j},\omega_{\{-i,j\}},0}},\mathbb{\mu}^{\pi,t-1}_{f_{\omega_{-j},\omega_{\{-i,j\}},1}} \right)\right)
\end{split} 
\end{equation*}
In total we have,
\begin{equation*}
\begin{split}
\sup_{f \in  C} R_{\pi} 
&\geq \frac{M_5}{2} \frac{1}{2^{d_y M^{d_x}} } \frac{s^2 (1-4s)^d}{2 M^{d_x+2}} T \sum_{j=1}^{M^{d_x}} \sum_{\omega_{-j}} \sum_{i=1}^{d_y} \sum_{\omega_{(-i,j)}} \exp\left(-z_{\{\omega_{-j},\omega_{-i,j}\}}\right)\\
&= \frac{M_5}{2} \frac{1}{2^{d_y M^{d_x}}} 2^{-(7+d_x)} M^{-d_x-2} T \sum_{j=1}^{M^{d_x}} \sum_{\omega_{-j}} \sum_{i=1}^{d_y} \sum_{\omega_{(-i,j)}} \exp\left(-z_{\{\omega_{-j},\omega_{-i,j}\}}\right)\\
\end{split} 
\end{equation*}
The last equality follows by setting $s=1/8$.

Combining these 2 bounds together, 
\begin{equation*}
\sup_{f \in C} R_{\pi} \geq \frac{1}{2^{d_y M^{d_x}}} \sum_{j=1}^{M^{d_x}} \sum_{\omega_{-j}} \sum_{i=1}^{d_y} \sum_{\omega_{(-i,j)}} \left(\frac{c_1 T}{2^{d_x} M^{d_x+2}} \exp\left(-z_{\omega_{-j},\omega_{\{-i,-j\}}}\right)+c_2 M^2 z_{\omega_{-j},\omega_{\{-i,-j\}}}\right)
\end{equation*}
To find a $z$ minimizing the right side, we use the first order condition.
\begin{equation*}
-\dfrac{c_1 T}{2^{d_x} M^{d_x+2}} \exp(-z)+c_2 M^2=0
\end{equation*}
Then,
\begin{equation*}
z^*=\log\left(\frac{c_1 T}{c_2 2^{d_x} M^{d_x+4}}\right)
\end{equation*}
Replacing $z$ by $z^*$ in the RHS, 
\begin{equation*}
\sup_{f \in C} R_{\pi} \geq M^{d_x+2} \left[1+\log\left(\dfrac{c_1 T}{c_2 2^{d_x} M^{d_x+4}}\right)\right]
\end{equation*}
Applying the first order condition again, we have 
\begin{equation*}
M^*= T^{\frac{1}{d_x+4}} 2^{-\frac{d_x}{(d_x+2)(d_x+4)}}
\end{equation*}
And finally,
\begin{equation*}
\sup_{f \in C} R_{\pi} \geq O(T^{\frac{d_x+2}{d_x+4}})
\end{equation*}

\end{document}